\documentclass[sigconf]{acmart}
\AtBeginDocument{%
  \providecommand\BibTeX{{%
    \normalfont B\kern-0.5em{\scshape i\kern-0.25em b}\kern-0.8em\TeX}}}

\usepackage{url}
\usepackage{graphicx}
\usepackage{amsmath}

\usepackage{algorithm}
\usepackage[noend]{algpseudocode}
\usepackage{booktabs}
\usepackage{microtype}      
\usepackage{multirow}
\usepackage{amsfonts} 
\usepackage{url} 
\usepackage{ulem}
\usepackage{latexsym}

\copyrightyear{2021}
\acmYear{2021}
\setcopyright{acmcopyright}\acmConference[MM '21]{Proceedings of the 29th ACM International Conference on Multimedia}{October 20--24, 2021}{Virtual Event, China}
\acmBooktitle{Proceedings of the 29th ACM International Conference on Multimedia (MM '21), October 20--24, 2021, Virtual Event, China}
\acmPrice{15.00}
\acmDOI{10.1145/3474085.3475637}
\acmISBN{978-1-4503-8651-7/21/10}
\begin{document}

\renewcommand{\algorithmicrequire}{\textbf{Input:}}  
\renewcommand{\algorithmicensure}{\textbf{Output:}} 
\begin{abstract}
Inspired by the success of BERT, several multimodal representation learning approaches have been proposed that jointly represent image and text. These approaches achieve superior performance by capturing high-level semantic information from large-scale multimodal pretraining. In particular, LXMERT and UNITER adopt visual region feature regression and label classification as pretext tasks. However, they tend to suffer from the problems of noisy labels and sparse semantic annotations, based on the visual features having been pretrained on a crowdsourced dataset with limited and inconsistent semantic labeling.
To overcome these issues, we propose unbiased Dense Contrastive Visual-Linguistic Pretraining (DCVLP), which replaces the region regression and classification with cross-modality region contrastive learning that requires no annotations. 
Two data augmentation strategies (Mask Perturbation and Intra-/Inter-Adversarial Perturbation) are developed to improve the quality of negative samples used in contrastive learning.
Overall, DCVLP allows cross-modality dense region contrastive learning in a self-supervised setting independent of any object annotations. We compare our method against prior visual-linguistic pretraining frameworks
to validate the superiority of dense contrastive learning on multimodal representation learning. 
\end{abstract}

\title{Dense Contrastive Visual-Linguistic Pretraining} 

\author{Lei Shi$^{1}$, Kai Shuang$^{1}$, Shijie Geng$^{2}$, Peng Gao$^{3}$, Zuohui Fu$^{2}$, Gerard de Melo$^{4}$\\Yunpeng Chen$^{5}$, Sen Su$^{1}$}
\affiliation{
  \institution{$^{1}$Beijing University of Posts and Telecommunications \country{China} \\ $^{2}$ Rutgers University\country{USA}\\ $^{3}$ Shanghai AI Laboratory\country{China} \\$^{4}$Hasso Plattner Institute, University of Potsdam\country{Germany} \\ $^{5}$YITU Technology\country{China}}
}
\email{{SLei, shuangk, susen}@bupt.edu.cn, {sg1309, zuohui.fu}@rutgers.edu}
\email{gaopeng@pjlab.org.cn,gdm@demelo.org, yunpeng.chen@yitu-inc.com}

\makeatletter
\def\authornotetext#1{
 \if@ACM@anonymous\else
 \g@addto@macro\@authornotes{
  \stepcounter{footnote}\footnotetext{#1}}
 \fi}
\makeatother
\authornotetext{Work done during an internship at Yitu Tech.}
\authornotetext{Corresponding author.}








\fancyhead{}





\maketitle

\section{Introduction}
Large-scale self-supervised pretraining of Transformers~\cite{devlin2018bert,radford2018improving} has had a transformative effect on Natural Language Processing (NLP). Models such as BERT \cite{devlin2018bert} and RoBERTa \cite{liu2019roberta} have led to strong gains across a wide swath of diverse NLP tasks. Inspired by this, Visual-Linguistic Pretraining (VLP) has been proposed to learn  multimodal models covering both vision and language, by adding extra masked prediction self-supervised strategies for the visual branch~\cite{tan2019lxmert,lu2019vilbert}.

Among recent VLP approaches, LXMERT~\cite{tan2019lxmert} and UNITER~\cite{chen2019uniter} are two prominent representatives. They perform feature regression or classification of masked visual regions as the pretext task of self-supervised learning.
\begin{figure}[t!]
 \centering
 \includegraphics[width=0.95\columnwidth]{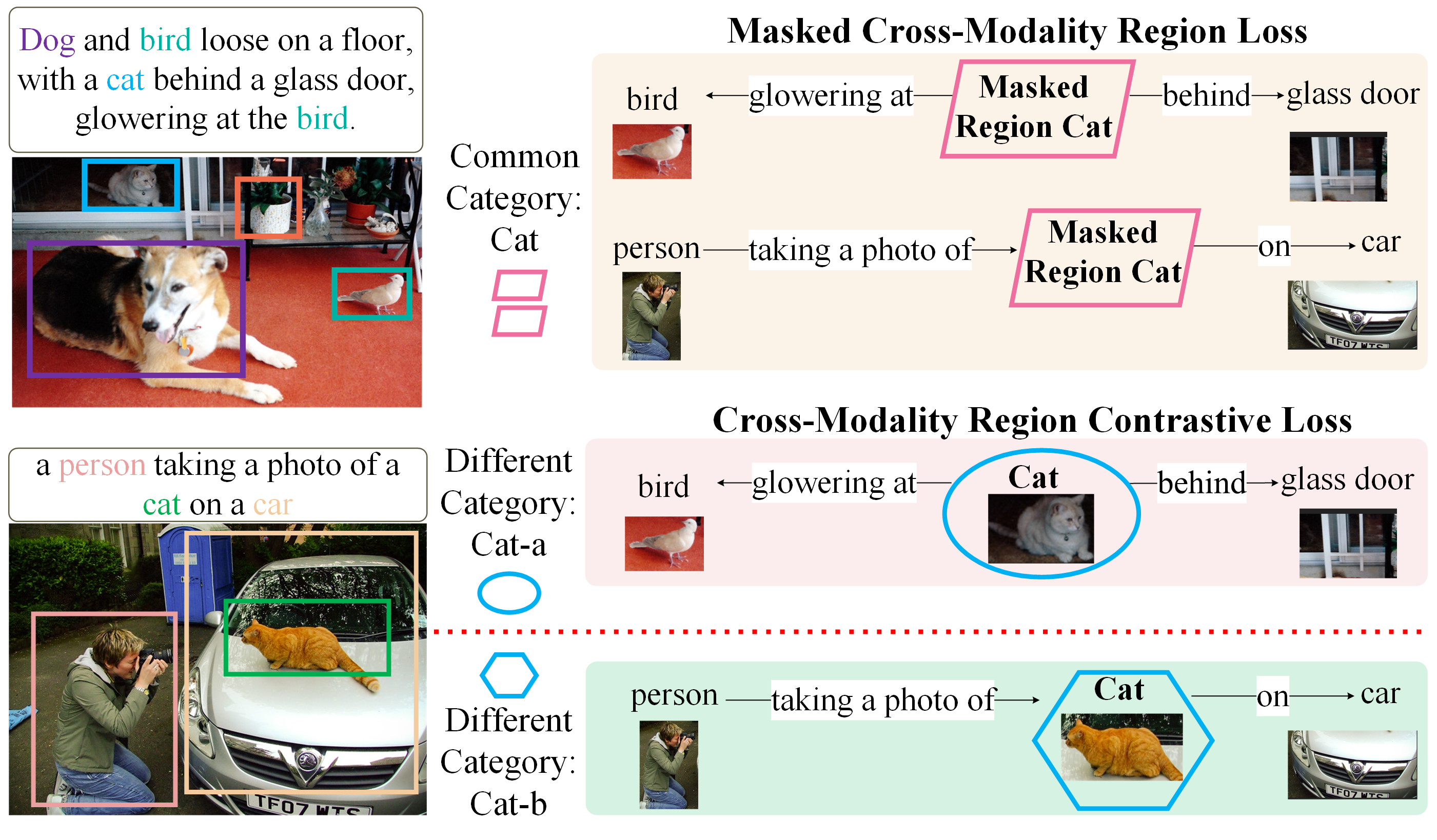}
 \caption{Different instances belonging to the same category will have different descriptions when in different scenes. However, the masked cross-modality region loss will classify them into the same category. Using the cross-modality region contrastive loss  can help our model to distinguish different instances in the same category because it can fuse pertinent semantic information in different scenarios.}
\label{fig:interact}
\vspace{2pt}
\end{figure}

However, we have identified several important drawbacks:
(i) Noisy Labels: $L_2$ feature regression and classification suffer from  noisy annotations in the Visual Genome dataset~\cite{krishna2017visual} used to train these models. The visual area labels in the data come from Faster-RCNN, which generally has relatively low confidence, unlike text labels.
(ii) Sparse Semantic Annotations: After establishing a relationship with the corresponding text caption information, it is often difficult to classify the visual region into a sparse label annotation. Especially when two different instances of the same category have divergent semantic information, a single classification is unable to capture the rich semantics, resulting in misleading gradients, as we can see in Figure~\ref{fig:interact}. In addition, previous approaches use masked region labels for a training loss, which is not sufficiently robust with respect to domain shifts. For instance, region labels annotated on MSCOCO \cite{lin2014microsoft} and VG \cite{krishna2017visual} have a large domain gap to NLVR2 \cite{suhr2017corpus}, which consists of images collected online and thus has entirely different image manifolds compared with the sorts of images used for pretraining.

To overcome the aforementioned noisy label and sparse semantic annotation problems, we propose  Dense Contrastive Visual-Linguistic Pretraining (DCVLP). 
 
Specifically, DCVLP replaces the region regression and classification with contrastive learning, which simultaneously avoids both of the above problems. 
Contrastive learning~\cite{hadsell2006dimensionality} aims to discriminate between positive examples and negative ones, which does not require any annotation and thus sidesteps the noisy label and sparse semantic annotation issues. However, the strength of such comparative learning hinges on our ability to identify high-quality negative samples. In vision--language pre-training, the input visual features directly come from bottom-up features~\cite{anderson2018bottom}, so we cannot directly use the rotation, filter, and resize operations on the original images to generate positive and negative samples~\cite{chen2020simple}. At the same time, classic forms of data augmentations in the field of visual contrastive learning tend to focus on natural attributes of the image itself. Thus, cross-modal visual language pre-training requires new data augmentation methods that can reflect high-level semantic changes in image context.
To this end, we explore two new methods for generating positive and negative sample pairs, namely mask perturbation and intra-/inter-adversarial perturbation for vision region contrastive learning. Mask perturbation can make positive sample pairs remain similar in the absence of regions or texts. Intra-/inter-adversarial perturbation is able to make the positive sample pairs as similar as possible under the interference category.
Our main contributions can be summarized as below:
\begin{itemize}
\item We propose a novel dense contrastive learning framework for visual-linguistic pretraining that avoids the sparse semantic annotation and noisy label problems encountered by previous visual-linguistic pretraining approaches such as LXMERT and UNITER. 
\item In order to discover fine-grained relationships between image objects and text descriptions, we first explore two new data augmentation methods, mask perturbation and intra-/inter-adversarial perturbation, to facilitate contrastive visual-linguistic pretraining.
\item We carry out an extensive set of empirical studies over variants of DCVLP to validate our proposed approach. Our DCVLP pretraining achieves significant improvements over strong baselines (LXMERT, UNITER) on downstream visual-linguistic tasks, especially when there is a domain gap between the pretraining and finetuning stages.
\end{itemize}

\begin{figure*}[htbp]
 \centering
 \includegraphics[width=0.86\textwidth]{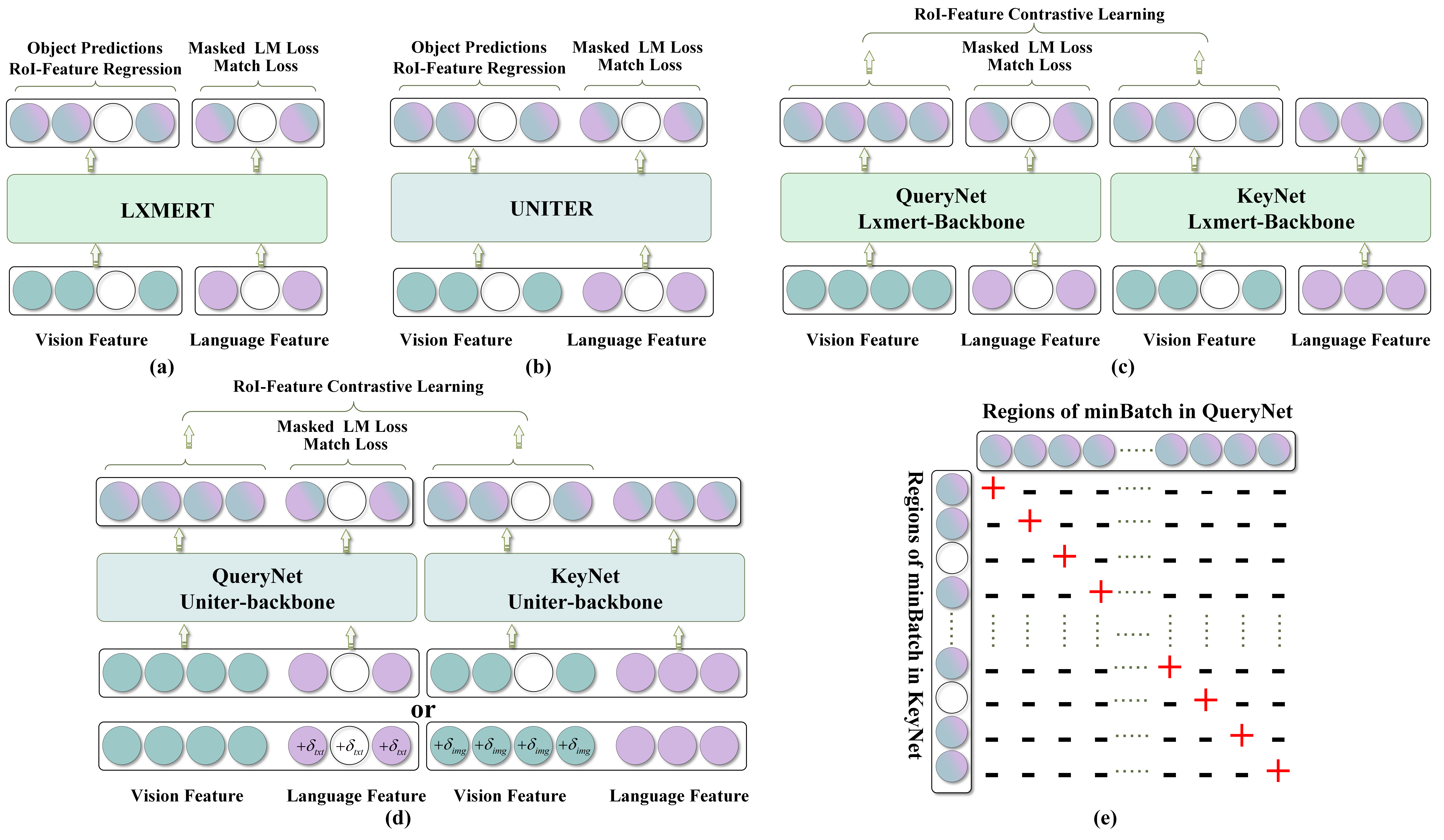}
 \caption{(a) represents the LXMERT model, which has dual-stream encoders.
The visual and linguistic features are first processed in independent encoders and then use a cross-modal Transformer for multimodal information fusion. (b) stands for the UNITER model, a single-stream encoder, i.e., visual and linguistic features are directly cascaded and then encoded.
 (c), (d) describe the overall architecture of the proposed DCVLP approach with LXMERT-backbone and UNITER-backbone, respectively; The masked token is represented by white circles. DCVLP includes a Query Network and a Key Network. The entire model is trained with a combination of three cross-modality losses. Cross-Modality Contrastive loss is applied to all regions. $\delta_{\rm{img}}$, $\delta_{\rm{txt}}$ in (d) denote adversarial perturbation. In (e), Regions in the same image position from QueryNet and KeyNet serve as positive pairs, while other regions are used to construct negative pairs. Since an image contains multiple object regions, the multiple positive pairs are from different image positions.} 
 \label{global2}
\end{figure*}

\section{Related Work}
\noindent\textbf{Self-supervised Learning in Vision and Language.~}
Deep Neural Networks (DNN) trained on ImageNet~\cite{deng2009imagenet} have revolutionized automatic feature representation learning~\cite{krizhevsky2012imagenet}. Compared to supervised training, which incurs a substantial cost for data annotation, self-supervised learning learns useful features automatically by constructing a loss from a pretext task, which does not require human annotation. 
In computer vision, context encoders~\cite{pathak2016context} learn features by image in-painting. Jigsaw~\cite{noroozi2016unsupervised} learns features by predicting the position of permuted features. Kolesnikov et al.~\cite{kolesnikov2019revisiting} carry out a large-scale study of previously proposed self-supervised learning methods and show that the performance of self-supervised tasks varies as the backbone changes. In NLP, large-scale pretraining with next-word prediction (GPT)~\cite{radford2018improving}, next sentence prediction, or masked word prediction (BERT)~\cite{devlin2018bert}, typically trained with the Transformer architecture~\cite{vaswani2017attention}, is now the predominant approach for state-of-the-art results in many NLP tasks.
Motivated by the success of transformer-based self-supervised learning in both vision and language, DFAF~\cite{gao2019dynamic}, QBN~\cite{shi2020multi}, MCAN~\cite{yu2019deep} and STSGR~\cite{geng2021dynamic} has shown that transformer can perform efficient multi-modality fusion. UNITER~\cite{chen2019uniter} and LXMERT~\cite{tan2019lxmert} have shown that masked words and visual regions can yield a good visual-linguistic representation. 

\vspace*{2mm}
\noindent\textbf{Contrastive Learning \& Adversarial Learning.~}
Contrastive learning is a sub-branch of self-supervised learning, employing a contrastive loss to learn a representation that is useful in downstream tasks.
The contrastive loss encourages the encoded instance features to be similar to positive keys while keeping away from negative ones.
Different contrastive learning methods adopt different strategies to generate positive and negative keys, which is an essential factor for the quality of the learned representation.
Wu et al.~\cite{wu2018unsupervised} select the keys from a large memory bank that stores the instance features for the entire training dataset.
Some works~\cite{tian2019contrastive,chen2020simple} generate keys using the current mini-batch samples.
MoCo V1, V2~\cite{he2019momentum,chen2020improved} proposes a momentum encoder to generate the keys on-the-fly and store them in a fixed-size queue. However, in each iteration, only part of the information in the negative sample queue is updated. Different from MoCo, our dense contrastive learning uses the original form of contrastive learning without utilizing a memory bank. 
BYOL~\cite{grill2020bootstrap} no longer uses negative samples, but rather attempts to learn using only positive samples. AdCo~\cite{hu2020adco} generates negative samples directly through training.
Recently, CLIP~\cite{radford2021learning} has successfully trained visual-linguistic representation using contrastive learning. Different from CLIP which simply utilizes global information during pretraining, our DCVLP employs dense contrastive learning which can better model fine-grained information. Neural networks are susceptible to adversarial attacks in the form of small perturbations~\cite{szegedy2013intriguing,goodfellow2014explaining,moosavi2017universal}. Adversarial learning aims at improving the robustness of neural network models to defend against adversarial attacks. Madry et al.~\cite{madry2018towards} conducts effective adversarial training through  Projected Gradient Descent (PGD). Shafahi et al.~\cite{shafahi2019adversarial} proposes to make neural network more robust compared with naively trained neural networks by updating adversarial perturbations.
FreeLB~\cite{zhu2020freelb} introduces adversarial training to improve the robustness and generalization of pretrained language models on NLP tasks. Recently, VILLA~\cite{gan2020large} proposed a similar adversarial training strategy to benefit multimodal representation learning.

\vspace*{2mm}
\noindent\textbf{Semantically Dense Learning.~}
Key tasks in image understanding include object detection, scene graph decomposition and image captioning. Single objects can contain dense semantic information when multiple objects appear in the image and information can flow between them.
Recent research \cite{desai2020virtex,bulent2020learning,ren2020scene} investigates this aspect and shows that image captioning annotations allow us to learn better image representations than simple object category labels. 
Such labels also benefit several downstream tasks such as object detection and semantic segmentation with 10 times fewer annotations. In the VQA task~\cite{li2019relation}, a novel implicit and explicit relation graph has been introduced to encode dense semantic information into a single object vector. Our proposed DCVLP framework can also infuse dense information into region and word vectors by drawing on contrastive learning. 

\section{Dense Contrastive Visual-Linguistic Pretraining}
Figure~\ref{global2} succintly summarizes the original LXMERT and UNITER models in (a)--(b), the structure of the DCVLP model in (c)--(d), and the composition of the dense contrast loss in (e).
As illustrated in Figure~\ref{global2}, (c)--(d), the architecture of DCVLP consists of a Query Network (QueryNet) and a Key Network (KeyNet). For LXMERT, they both contain a language Transformer encoder, a vision Transformer encoder and a multi-modal fusion Transformer. For UNITER, they only contain a single multi-modal self-attention Transformer encoder. LXMERT is a dual stream model, while UNITER is a single stream model. Our DCVLP is composed of QueryNet and KeyNet. We use either the LXMERT model or the UNITER model as the respective backbone of DCVLP in Figure~\ref{global2}, (c)--(d). KeyNet is copied from QueryNet with the same layers and parameters. In Figure~\ref{global2}, (a)--(b), both LXMERT and UNITER use mask region object predictions and mask region feature regression in the visual branch, which makes them susceptible to the  influence of noisy labels. Thus, we use a new RoI-Feature Contrastive Loss to replace the original loss function of the visual branch in LXMERT and UNITER, and retain the classic Cross-Modality Match Loss, Masked Language Model Loss as later given in Eqn.~(\ref{equ:7}) and (\ref{equ:6}).

\subsection{Multi-modality Fusion}
\label{sec:fusion}
Given image--sentence pairs from a vision--language dataset, we first tokenize each sentence using the WordPieces technique~\cite{wu2016google} and map a token $W_j$ to its corresponding embedding $h_{\rm{emb}}(W_j) \in {\mathbb{R}^{d_w}}$, where $d_w=768$. In addition, visual regions $B \in {\mathbb{R}^{N \times 4}}$ and their features $F \in {\mathbb{R}^{N \times d_o}}$ are extracted by a Faster-RNN~\cite{ren2015faster} detector pretrained on Visual Genome~\cite{krishna2017visual} for each image $I$: $B, F = \mathrm{RCNN}(I)$, where we detect $N = 36$ regions in the LXMERT-backbone (or $N = 10\sim100$ in UNITER-backbone) within each image and each such region is represented using a feature dimensionality of $d_o=2048$. Then we can calculate the visual inputs $v_i$ and textual inputs $w_j$ of DCVLP as follows:
\begin{equation}
\label{init-v-feat-init-w-feat}
 \begin{split}
     &{v_i} = \frac{{{g_{\rm{F}}}\left( {{F_i}} \right) + {g_{\rm{P - ROI}}}\left( {{B_i}} \right)}}{2}  \\
     &{w_j} = {h_{\rm{emb}}\left( {{W_j}} \right) + {h_{\rm{P - word}}}\left(P_j \right)}
  \end{split}
\end{equation}
where $g_{\rm{F}}$ and $g_{\rm{P - ROI}}$ are two fully-connected layers that map $F_i$ and $B_i$, respectively, to the feature dimensionality $d_w$, while $h_{\rm{P-word}}$ is a positional encoding function for the position $P_j$ of token $W_j$. 

Taking $v_i$ and $w_j$ as inputs, with mask perturbation, DCVLP adopts symmetrical masking for both QueryNet and KeyNet in Figure~\ref{global2}, (c)--(d). For the text modality, we uniformly choose 15\% of the input textual tokens for replacement. The chosen tokens are replaced by the special \textit{[MASK]} token. For the visual modality, we use a different masking strategy: the features of the chosen regions can either be set to zero or be replaced by region features from other images.

DCVLP contains two networks called QueryNet and KeyNet. QueryNet and KeyNet have exactly the same network structure. In the experiment, we can use the LXMERT or UNITER model as the backbone for QueryNet and KeyNet.

With LXMERT backbone for DCVLP in Figure~\ref{global2}-(c) the three encoders are implemented by 3 modules, namely, the visual self-attention, language self-attention and visual-linguistic co-attention modules. Visual self-attention performs information fusion between region features by using such features as both key, query and value in the attention model. We denote the key, query and value features for visual as $K_v$, $Q_v$, $V_v$, and for language as $K_w$, $Q_w$, $V_w$, respectively. Then the intra-modality information fusion for visual and language features can be denoted as:
\begin{equation}\label{eq:array}
\begin{split}
     &\widehat v = \rm{Intra}_{v \leftarrow v}\left( {{Q_v},{K_v},{V_v}} \right)\\
     &\widehat w{\rm{ }} = \rm{Intra}_{{w \leftarrow w}}\left( {{Q_w},{K_w},{V_w}} \right)
\end{split}
\end{equation}
where the attention module of a Transformer layer can be expressed as follows:
\begin{equation}
    \rm{Attention}(Q,K,V) = \mathrm{Softmax} (Q{K^T}/\sqrt d )V
\end{equation}
After deploying intra-modality information flow for language and visual signals, we invoke an inter-modality fusion module to connect the signals in the language and visual features. The inter-modality fusion process is bi-directional, which includes information fusion from language to vision and vice versa:
\begin{equation}
\label{coattention-w-v}
\begin{split}
     &\widetilde v = \rm{Inter}_{v \leftarrow w}\left( {{Q_v},{K_w},{V_w}} \right) \\
     &\widetilde w = \rm{Inter}_{{w \leftarrow v}}\left( {{Q_w},{K_v},{V_v}} \right)
\end{split}
\end{equation}

After intra-inter modality feature fusion, we can acquire a multi-modality contextual feature embedding for each word and visual region. A contextual feature encodes the multimodal interactions in a compact feature vector. The contextual features are used by DCVLP for Modality Match Loss and Masked Cross-Modality Language Model Loss in the language branch and the RoI-Feature Contrastive Loss in the visual branch.
\par With UNITER-backbone for DCVLP, unlike LXMERT, UNITER is a single-stream model. Visual information and text information are cascaded, and then encoded by the self-attention module. Visual and language features can be denoted as:

\begin{equation}
\label{selfattention-wv}
\begin{split}
     &\widetilde v = \rm{Intra}_{v\leftarrow{v+w}} \left( 
     Q_{v+w}, K_{v+w}, V_{v+w} \right) \\
     &\widetilde w = \rm{Intra}_{w\leftarrow{v+w}} \left( Q_{v+w},K_{v+w},V_{v+w} \right)
\end{split}
\end{equation}

\subsection{Cross-Modality Match Loss, Masked Language Model Loss for Language Branch}
\label{sec:lang}
In the pretraining stage, DCVLP performs different pretext tasks compared with LXMERT and UNITER. DCVLP does not contain a supervised learning task and thus is independent of human-annotated labels. For the language branch, we keep masked language modeling and image--sentence matching prediction as two pretext tasks. The mask loss was first proposed by BERT \cite{devlin2018bert}. Subsequent visual-linguistic BERT approaches \cite{lu2019vilbert,li2019visualbert,lu2019vilbert,tan2019lxmert,chen2019uniter,li2020oscar} add a visual feature mask loss besides the masked language modeling loss. This loss masks out the contextual representation obtained in Section~\ref{sec:fusion} and predicts the masked feature using its contextual information. By optimizing the mask loss, the Transformer implicitly learns to encode contextual information, which facilitates the generalization on downstream tasks. 
In DCVLP, we only apply this mask loss for the textual inputs.
Additionally, we also add a matching loss, which involves a binary Yes/No classification to predict whether the sentence matches the visual feature. The mask loss can be formalized as follows:
\begin{equation}
     {\mathcal{L}_{\rm{MLM}}} =  - {E_{w\sim D}}\log {P_\theta }\left( {{w_m}|\widetilde {{w_{/m}}}} \right),
\label{equ:6}
\end{equation}
where $\theta$ denotes the parameters of the Language Transformer Encoder, $w_m$ and $\widetilde {{w_{/m}}}$ denote the masked token to be predicted and the contextual tokens that are not masked. The matching loss is defined as:
\begin{equation}
 \begin{split}
     {\mathcal{L}_{\rm{MATCH}}}  = & - E_{w_{\rm{CLS}}\sim D} \big[ y\log P_\theta \big( \widetilde {w_{\rm CLS}} \big) \\
     & + \big( 1 - y \big)\log \big( 1 - P_\theta \big( \widetilde {w_{\rm CLS}} \big) \big],
 \end{split}
\label{equ:7}
\end{equation}

which is clearly a binary classification task. In the above equation, $\widetilde {w_{\rm CLS}}$ stands for the \textit{[CLS]} token, which encodes the visual-linguistic information for tackling the image--sentence matching pretext task.

\subsection{RoI-Feature Dense Contrastive Loss for Visual Branch}
\label{sec:vis}
Contrastive learning performs self-supervised representation learning by discriminating visually similar representation pairs from a group of negative ones. Given visual region features extracted by Faster-RCNN, we can obtain a positive query--key pair by feeding such features into both QueryNet and KeyNet. All region features from other images in the sampled batch and different positions serve as negative keys. Then we conduct contrastive learning by updating network weights to minimize the following loss:
\begin{equation}
     \mathcal{L}_{\rm{C}} =  - \log \frac{\exp \left( {{s^ + }/\tau } \right)}{\exp \left( {{s^ + }/\tau } \right)  + \sum\nolimits_{j = 0}^K {\exp \left( {s_j^ - /\tau } \right)}}
\end{equation}
\begin{equation}
\begin{split}
     & {s^ + } = \widetilde{v_i^{\rm query}} \cdot \widetilde{v_i^{\rm key+}}\\
     & {s^ - } = \widetilde{v_i^{\rm query}} \cdot \widetilde{v_j^{\rm key-}}
\end{split}
\label{equ:8-9}
\end{equation}

Here, $\tau$ is the temperature of the softmax operation, $\widetilde{v_i^{\rm key+ }}$ is a positive key of $\widetilde{v_i^{\rm query}}$, and $\widetilde{v_j^{\rm key-}}$ serves as negative examples for calculating ${\mathcal{L}_{\rm{CONTRAST}}}$ $({\mathcal{L}_{\rm{C}}})$.
In order to generate positive sample pairs, one simple strategy is to still use the mask operation, and another is to incorporate adversarial perturbation based on the PGD algorithm~\cite{madry2018towards}. In the following, we outline these two methods.

\subsubsection{Mask Perturbation Strategy for Visual Branch Dense Contrast Learning}
\label{mask perturbation}
In DCVLP, QueryNet and KeyNet have the same network structure, and we can use LXMERT or UNITER as the backbone for them. Specifically, in our experiments, we first consider a UNITER backbone as an example and finally also report the results of LXMERT as the backbone.
In the mask perturbation DCVLP model, for QueryNet, we use a mask operation for textual information, while the visual information does not use any masking. Symmetrically, in KeyNet, we use masking for visual information, while the textual signals are not masked. The mask operation method is the same as in the UNITER model. The two sets of multimodal information processed by the mask are fed into QueryNet and KeyNet for encoding, respectively, and then the visual information is obtained from the last layer of QueryNet and KeyNet, respectively. The visual region information at the same position of QueryNet and KeyNet forms a pair of positive examples and the region information at different positions forms a negative example. We can consider different mask strategies as shown in Table~\ref{tab:mask-sys}. We empirically found that it is better to use the symmetric mask strategy for the multimodal information in QueryNet and KeyNet.

\begin{table}[]
\scalebox{0.92}{
\begin{tabular}{llllccc}
\hline
\multicolumn{2}{c}{}       & QueryNet & KeyNet   &                   & \begin{tabular}[c]{@{}c@{}}VQA\\ (Test\\ dev)\end{tabular} & \begin{tabular}[c]{@{}c@{}}NLVR2\\ (Test\\ -P)\end{tabular} \\ \hline
\multirow{2}{*}{1} & Text  & Mask\_1  & Mask\_1  & \multirow{2}{*}{} & \multirow{2}{*}{72.45}                                     & \multirow{2}{*}{77.30}                                      \\
                   & Image & Mask\_a  & Mask\_a  &                   &                                                            &                                                             \\ \hline
\multirow{2}{*}{2} & Text  & Mask\_1  & Mask\_1  & \multirow{2}{*}{} & \multirow{2}{*}{73.30}                                     & \multirow{2}{*}{77.57}                                      \\
                   & Image & No\_Mask & Mask\_a  &                   &                                                            &                                                             \\ \hline
\multirow{2}{*}{3} & Text  & Mask\_1  & Mask\_2  & \multirow{2}{*}{} & \multirow{2}{*}{73.14}                                     & \multirow{2}{*}{78.45}                                      \\
                   & Image & No\_Mask & Mask\_a  &                   &                                                            &                                                             \\ \hline
\multirow{2}{*}{4} & Text  & Mask\_1  & Mask\_2  & \multirow{2}{*}{} & \multirow{2}{*}{72.30}                                     & \multirow{2}{*}{77.18}                                      \\
                   & Image & Mask\_b  & Mask\_a  &                   &                                                            &                                                             \\ \hline
\multirow{2}{*}{5} & Text  & Mask\_1  & No\_mask & \multirow{2}{*}{} & \multirow{2}{*}{73.54}                                     & \multirow{2}{*}{78.70}                                      \\
                   & Image & No\_mask & Mask\_a  &                   &                                                            &                                                             \\ \hline
\end{tabular}
}
\caption{QueryNet and KeyNet apply UNITER-backbone. For text modality, Mask-1, Mask-2 indicate that two Mask operations are performed on the text description information, and the tokens masked are different each time. For the image region feature, Mask-a, Mask-b indicate that two Mask operations are performed on the region contained in a picture, and the region masked is different each time. No-mask represents the use of original visual or textual information.}
\label{tab:mask-sys}
\end{table}

\subsubsection{Intra Adversarial Perturbation Strategy For Visual Branch Dense Contrast Learning}
\label{adversarial perturbation}
We use the PGD attack algorithm to induce adversarial perturbation $\delta_{\rm{img}}$, $\delta_{\rm{txt}}$ for the visual region feature space and text embedding feature space. In other words, we aim to use the idea of contrastive learning to make the tokens in the same position as similar as possible when the adversarial perturbations $\delta_{\rm{img}}$, $\delta_{\rm{txt}}$ are injected. The intra-adversarial perturbation strategy generates $\delta_{\rm{img}}$, $\delta_{\rm{txt}}$ that directly maximize $\mathcal{L}_{\rm{C}}$. Similar to the above mask strategy idea, we add adversarial perturbation symmetrically to QueryNet and KeyNet. The algorithmic details are formalized in Algorithm~\ref{alg::acl}, where $\mathcal{L}_{\rm{C}}$ follows Eqn.~(\ref{equ:8-9}). Symmetrically injecting adversarial perturbation in QueryNet and KeyNet, respectively, can alleviate the degree of simultaneous attacks on the network~\cite{jiang2020robust}, which is naturally suitable for multimodal conditions. In the Intra-Adversarial Perturbation Strategy For Visual Branch Dense Contrast Learning (DCVLP\_ADV-intra), we first use gradient ascent with regard to $\mathcal{L}_{\rm{C}}$ to generate adversarial perturbation in the adversarial training process, which may easily lead to inconsistent representations of tokens in the same position of QueryNet and KeyNet. With multiple gradient ascent steps, we use the contrastive loss to minimize dissimilarities of the tokens in the same position. This process can track changes in the global sampled-batch signal across time and produce negative samples that are more difficult to distinguish.

\subsubsection{Inter-Adversarial Perturbation Strategy For Visual Branch Dense Contrast Learning}
\label{inter adversarial perturbation} The Intra-Adversarial Perturbation Strategy is a very direct method~\cite{jiang2020robust,hu2020adco}, but its effect is found to be suboptimal in our experiments. As using unsupervised $\mathcal{L}_{\rm{C}}$ without a clear category for guidance to optimize $\delta_{\rm{img}}$, $\delta_{\rm{txt}}$ is very difficult, it is hard for $\mathcal{L}_{\rm{C}}$ to drive positive sample pairs to be similar with such attacks. Thus, we propose to instead use $\mathcal{L}_{\rm{MLM}}$ to generate $\delta_{\rm{img}}$, $\delta_{\rm{txt}}$. The algorithmic details are given in  Algorithm~\ref{alg::acl_inter}. This form of perturbation attempts to increase the gradient of the masked language model, but we minimize $\mathcal{L}_{\rm{C}}$ to defend against such perturbation. Our approach is different from VILLA~\cite{gan2020large} because VILLA uses the same loss when generating perturbations and when defending against perturbation. However, that method will rely too much on the labels, and especially when there is an error or bias in the labels, the approach will not work well. Hence, we propose Inter-Adversarial Perturbation Strategy For Visual Branch Dense Contrast Learning (DCVLP\_ADV-inter), which combines $\mathcal{L}_{\rm{MLM}}$ and $\mathcal{L}_{\rm{C}}$ to generate negative samples with moderate difficulty. It does not rely on any labels and at the same time promotes the joint optimization of both $\mathcal{L}_{\rm{MLM}}$ and $\mathcal{L}_{\rm{C}}$.

\begin{figure*}[htp!]
 \centering
 \includegraphics[width=0.9\textwidth]{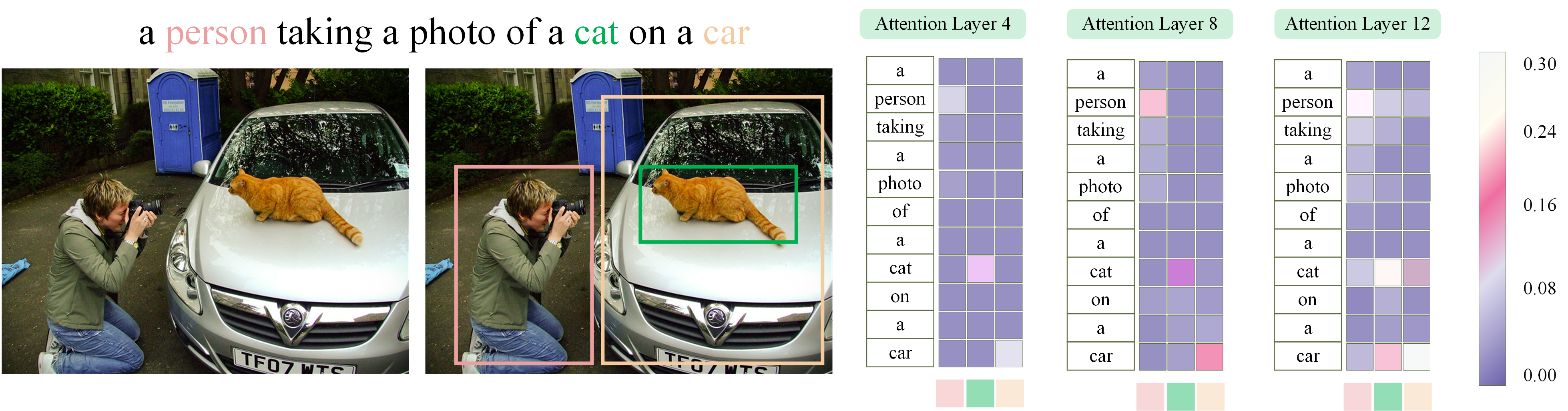}
 \caption{Illustration of attention weights in DCVLP with UNITER-backbone. The pink, green and orange in the right represents the people, cat and car bounding boxes in the left figure correspondingly. The lighter the color, the greater the weight of attention.}
 \label{vision}
\end{figure*}
\begin{algorithm}[h]
  \caption{Multimodal Intra-Adversarial Training used in DCVLP}
  \label{alg::acl}
  \begin{algorithmic}[1]
    \Require
      $\textbf{\textit{v}}$,\,$\textbf{\textit{w}}$:\,clean image and text in sampled-batch;\,$\delta_{\rm{img}}$,\, $\delta_{\rm{txt}}$: adversarial perturbation in image and text; \,$\epsilon$:\, perturbation bound;\, $q$,\,$k$ represent $\textit{QueryNet}$ and $\textit{KeyNet}$ with parameters $\theta$;\, $\textit{K}$: ascent steps ;\,$\mathcal{L}_{\rm{C}}$ represent contrast loss.
    \Ensure
      Parameters $\theta$ in $\textit{QueryNet}$ and $\textit{KeyNet}$;
       \While {$t < \rm{Iter}_{\max}$}
         \For{$i=1...\textit{K}$}
      \State Accumulate gradient of $\theta$ given $\delta_{\rm{img},\textit{$i-1$}}$ and $\delta_{\rm{txt},\textit{$i-1$}}$
      
      \State {$\delta_{\rm{img}}$,\, $\delta_{\rm{txt}} = \mathop{\arg\max}\limits_{{{{\left\| {{\delta_{\rm{img}}}} \right\|}_\infty } \le \varepsilon ,{{\left\| {{\delta_{\rm{txt}}}} \right\|}_\infty } \le \varepsilon }} \left\langle\mathcal{L}_{\rm{C}}\right\rangle$
      \State $\mathcal{L}_{\rm{C}}$ $\overset{\rm{def}}{=}$ ${q( {v,w + {\delta_{\rm{txt}}};\theta } ) \circ k( {v + {\delta_{\rm{img}}},w;\theta } )}$}
      \State Update $\delta_{\rm{img}}$,\, $\delta_{\rm{txt}}$ via gradient ascent
      
     \EndFor
      \State Update parameters $\theta$ to minimize $\mathcal{L}_{\rm{C}}$;
    \EndWhile
  \end{algorithmic}
\end{algorithm}

\begin{algorithm}[h]
  \caption{Multimodal Inter-Adversarial Training used in DCVLP}
  \label{alg::acl_inter}
  \begin{algorithmic}[1]
    \Require
      $\textbf{\textit{v}}$,\,$\textbf{\textit{w}}$:\,clean image and text in sampled-batch;\,$\delta_{\rm{img}}$,\, $\delta_{\rm{txt}}$: adversarial perturbation in image and text; \,$\epsilon$:\, perturbation bound;\, $q$,\,$k$ represent $\textit{QueryNet}$ and $\textit{KeyNet}$ with parameters $\theta$;\, $\textit{K}$: ascent steps ;\,$\mathcal{L}_{\rm{C}}$ represents contrast loss:\, $\mathcal{L}_{\rm{MLM}}$ represents masked language model loss.
    \Ensure
      Parameters $\theta$ in $\textit{QueryNet}$ and $\textit{KeyNet}$;
       \While {$t < \rm{Iter}_{\max}$}
         \For{$i=1...\textit{K}$}
      \State Accumulate gradient of $\theta$ given $\delta_{\rm{img},\textit{$i-1$}}$ and $\delta_{\rm{txt},\textit{$i-1$}}$
      
      \State {$\delta_{\rm{img}}$,\, $\delta_{\rm{txt}} = \mathop{\arg\max}\limits_{{{{\left\| {{\delta_{\rm{img}}}} \right\|}_\infty } \le \varepsilon ,{{\left\| {{\delta_{\rm{txt}}}} \right\|}_\infty } \le \varepsilon }} \left\langle\mathcal{L}_{\rm{MLM}}\right\rangle$
      \State $\mathcal{L}_{\rm{C}}$ $\overset{\rm{def}}{=}$ ${q( {v,w + {\delta_{\rm{txt}}};\theta } ) \circ k( {v + {\delta_{\rm{img}}},w;\theta } )}$}
      \State Update $\delta_{\rm{img}}$,\, $\delta_{\rm{txt}}$ via gradient ascent
      
     \EndFor
      \State Update parameters $\theta$ to minimize $\mathcal{L}_{\rm{C}}$;
    \EndWhile
  \end{algorithmic}
\end{algorithm}

\section{Experiments}
In this section, we first introduce the implementation details of the proposed  Dense Contrastive Visual-Linguistic Pretraining framework. Then we conduct extensive comparative studies to demonstrate the effectiveness of the proposed method.
We apply the DCVLP algorithm to the LXMERT and UNITER frameworks, respectively. With LXMERT backbone for DCVLP, the pre-training data is the same as LXMERT, namely MSCOCO and Visual Genome. To assess the learned visual-linguistic features, we conduct finetuning experiments and compare LXMERT backbone for DCVLP against the original LXMERT on three downstream tasks, i.e., VQA v2~\cite{balanced_vqa_v2}, GQA~\cite{hudson2019gqa} and NLVR2~\cite{suhr2018corpus}, as considered in the LXMERT paper. 
With UNITER-backbone for DCVLP, the pre-training data is the same as for UNITER. The UNITER model uses a large-scale V+L dataset composed of four subsets: (i) COCO~\cite{lin2014microsoft}; (ii) Visual Genome (VG)~\cite{krishna2017visual} [21]; (iii) Conceptual Captions (CC)~\cite{sharma2018conceptual}; and (iv) SBU Captions~\cite{ordonez2011im2text}.
We conduct finetuning experiments and compare UNITER-backbone for DCVLP against the original UNITER on six downstream tasks, including: (i) VQA; (ii) Visual Commonsense Reasoning (VCR)~\cite{zellers2019recognition}; (iii) NLVR2; (iv) Visual Entailment~\cite{xie2019visual}; (v) Image--Text Retrieval (including zero-shot setting)~\cite{lee2018stacked}; and (vi) Referring Expression Comprehension~\cite{nagaraja2016modeling}. 

\begin{table*}[]
\scalebox{0.92}{
\begin{tabular}{lccccccccc}
\hline
\textbf{Method}                                                        & \multicolumn{2}{c}{\textbf{VQA}}      & \multicolumn{3}{c}{\textbf{VCR}}             & \textbf{GQA}   &\textbf{NLVR2}     &\textbf{SNLI-VE} \\ \cline{2-9} 
\textbf{}                                                              & \textbf{test-dev} & \textbf{test-std} & \textbf{Q-A} & \textbf{QA-R} & \textbf{Q-AR} & \textbf{dev}   & \textbf{test-P}   & \textbf{test}     \\ \hline
\textbf{Label Supervision}\\
ViLBERT                                                                & 70.55             & 70.92             & 72.42(73.3)  & 74.47(74.6)   & 54.04(54.8)   & -              & -                 & -                 \\
VisualBERT                                                             & 70.80             & 71.00             & 70.8(71.6)   & 73.2(73.2)    & 52.2(52.4)    & -              & 67.0              & -                 \\
LXMERT                                                                 & 72.42             & 72.54             & -            & -             & -             & 61.39          & 74.50             & -                 \\
Unicoder-VL                                                            & -                 & -                 & 72.6(73.4)   & 74.5(74.4)    & 54.4(54.9)    & -              & -                 & -                 \\
12-in-1                                                                & 73.15             & -                 & -            & -             & -             & -              & 78.87             & 76.95             \\
VL-BERT-Base                                                           & 71.16             & -                 & 73.8(-)      & 74.4(-)       & 55.2(-)       & -              & -                 & -                 \\
Oscar-Base                                                             & 73.16             & 73.44             & -            & -             & -             & 61.58          & 78.36             & -                 \\
VILLA-Pre                                                              & 73.03             & -                 & 74.76        & 77.04         & 57.82         & -              & 78.44             & 78.43             \\
UNITER-Base                                                            & 72.70             & 72.91             & 74.56(75.0)  & 77.03(77.2)   & 57.76(58.2)   & -              & 77.85             & 78.28             \\ \hline
\textbf{No Label Supervision}\\
\begin{tabular}[l]{@{}l@{}}DCVLP\_MASK\\ (LXMERT backbone)\end{tabular} & 73.07             & 73.15             & -            & -             & -             & 61.78          & 76.81             & -                 \\ \hline
\begin{tabular}[l]{@{}l@{}}DCVLP\_MASK\\ (UNITER backbone)\end{tabular} & 73.54             & 73.73             & 74.90(75.56) & 77.12(77.3)   & 57.94(58.4)   & -              & 78.70             & 78.70              \\
\begin{tabular}[l]{@{}l@{}}DCVLP\_ADV-intra\\ (UNITER-backbone)\end{tabular}  & 73.15             & 73.34             & 74.46(74.38) & 76.89(77.08)  & 57.59(58.10)   & -              & 78.13             & 78.34             \\
\begin{tabular}[l]{@{}l@{}}DCVLP\_ADV-inter\\ (UNITER-backbone)\end{tabular}  & 73.33             & 73.49             & 74.72(74.70) & 77.00(77.28)  & 57.70(58.3)   & -              & 78.43             & 78.61             \\ \hline
\end{tabular}
}
\caption{Results on VQA, VCR, GQA, NLVR2, and SNLI-VE; \_MASK and \_ADV-intra/inter denotes Mask and Intra/Inter Adversarial Perturbation Strategy respectively. }
\label{tab:main-1}
\end{table*}

\begin{table*}[]
\centering
\scalebox{0.92}{
\begin{tabular}{lcccccccccccc}
\hline
\multirow{2}{*}{\textbf{Method}}                                        & \multicolumn{6}{c}{\textbf{RefCOCO+}}                                                                                                                                                                                                     & \multicolumn{6}{c}{\textbf{RefCOCO}}                                                                                                                                                                                                      \\ \cline{2-13} 
                                                                        & \textbf{val} & \textbf{testA} & \textbf{testB} & \textbf{\begin{tabular}[c]{@{}c@{}}val\\ (d)\end{tabular}} & \textbf{\begin{tabular}[c]{@{}c@{}}testA\\ (d)\end{tabular}} & \textbf{\begin{tabular}[c]{@{}c@{}}testB\\ (d)\end{tabular}} & \textbf{val} & \textbf{testA} & \textbf{testB} & \textbf{\begin{tabular}[c]{@{}c@{}}val\\ (d)\end{tabular}} & \textbf{\begin{tabular}[c]{@{}c@{}}testA\\ (d)\end{tabular}} & \textbf{\begin{tabular}[c]{@{}c@{}}testB\\ (d)\end{tabular}} \\ \hline
\textbf{Label Supervision}\\
ViLBERT                                                                 & -            & -              & -              & 72.34                                                      & 78.52                                                        & 62.61                                                        & -            & -              & -              & -                                                          & -                                                            & -                                                            \\
VL-BERT                                                                 & 79.88        & 82.40          & 75.01          & 71.60                                                      & 77.72                                                        & 60.99                                                        & -            & -              & -              & -                                                          & -                                                            & -                                                            \\
VILLA-Pre                                                               & -            & -              & -              & -                                                          & -                                                            & -                                                            & -            & -              & -              & -                                                          & 87.34                                                        & 74.35                                                        \\
UNITER\_Base                                                          & 83.66        & 86.19          & 78.89          & 75.31                                                      & 81.30                                                        & 65.58                                                        & 91.64        & 92.26          & 90.46          & 81.24                                                      & 86.48                                                        & 73.94                                                        \\ \hline
\textbf{No Label Supervision}\\
\begin{tabular}[l]{@{}l@{}}DCVLP\_MASK\\ (UNITER-backbone)\end{tabular} & 83.83        & 86.69          & 79.26          & 75.45                                                      & 81.58                                                        & 65.82                                                        & 91.84        & 92.38          & 90.66          & 81.45                                                      & 86.85                                                        & 74.03                                                        \\
\begin{tabular}[l]{@{}l@{}}DCVLP\_ADV-intra\\ (UNITER-backbone)\end{tabular}    & 83.32       & 86.00          & 78.73          & 75.02                                                      & 81.15                                                        & 65.41                                                        & 91.33        & 92.02          & 90.21          & 80.96                                                      & 86.27                                                        & 73.50                                                        \\
\begin{tabular}[l]{@{}l@{}}DCVLP\_ADV-inter\\ (UNITER-backbone)\end{tabular}    & 83.77        & 86.63          & 79.01          & 75.23                                                      & 81.34                                                        & 65.76                                                        & 91.67        & 92.30          & 90.55          & 81.30                                                      & 86.65                                                        & 73.89                                                        \\ \hline
\end{tabular}
}
\caption{Results on RefCOCO+ and RefCOCO. The (d) denotes evaluation using detected proposals}
\label{tab:main-2}
\end{table*}

\begin{table*}[]
\centering
\scalebox{0.92}{
\begin{tabular}{lcccccccccc}
\hline
\multirow{2}{*}{\textbf{Method}}                                        & \multicolumn{4}{c}{\textbf{RefCOCOg}}                                                     & \multicolumn{3}{c}{\textbf{Flickr30 IR}}                           & \multicolumn{3}{c}{\textbf{Flickr30k TR}}                          \\ \cline{2-11} 
                                                                        & \textbf{val}         & \textbf{test}        & \textbf{val-d}       & \textbf{test-d}      & \textbf{R @ 1}       & \textbf{R @ 5}       & \textbf{R @ 10}      & \textbf{R @ 1}       & \textbf{R @ 5}       & \textbf{R @ 10}      \\ \hline
\textbf{Label Supervision}\\
ViLBERT                                                                 & -                    & -                    & -                    & -                    & 58.20                & 84.90                & 91.52                & -                    & -                    & -                    \\
Unicoder-VL                                                             & -                    & -                    & -                    & -                    & 71.50                & 90.90                & 94.90                & 86.20                & 96.30                & 99.00                \\
VILLA-Pre                                                               & -                    & -                    & -                    & -                    & 73.76                & 93.02                & 96.28                & -                    & -                    & -                    \\
UNITER\_Base                                                            & 86.52                & 86.52                & 74.31                & 74.51                & 72.52                & 92.36                & 96.08                & 85.90                & 97.10                & 98.80                \\ \hline
\textbf{No Label Supervision}\\
\begin{tabular}[l]{@{}l@{}}DCVLP\_MASK\\ (UNITER-backbone)\end{tabular} & 86.47                & 86.62                & 74.55                & 74.98                & 73.60                & 92.66                & 95.98                & 86.33                & 97.45                & 99.02                \\
\begin{tabular}[l]{@{}l@{}}DCVLP\_ADV-intra\\ (UNITER-backbone)\end{tabular}  & 86.48                & 86.13                & 74.10                & 74.33                & 73.01                & 92.08                & 95.49                & 85.50                & 96.87                & 98.43               \\
\begin{tabular}[l]{@{}l@{}}DCVLP\_ADV-inter\\ (UNITER-backbone)\end{tabular}  & 86.23                & 86.45                & 74.42                & 74.62                & 74.03                & 92.78                & 96.27                & 86.51                & 97.66                & 99.11                \\ \hline
\end{tabular}
}
\caption{Results on RefCOCOg and Flickr30k Image Retrieval (IR) and Text Retrieval (TR).}
\vspace*{-4mm}
\label{tab:main-3}
\end{table*}

\vspace*{2mm}
\noindent{\bf Implementation Details.~}
Following LXMERT and UNITER, we pretrain DCVLP on the same image--text pairs datasets from LXMERT and UNITER, respectively. In the pretraining stage, for a fair comparison, we retain the same basic parameters as LXMERT and UNITER respectively. For the dense contrastive loss for the visual branch in DCVLP, the temperature $\tau$ in the contrastive loss is set to 0.07. The embedding head dimensionality is 128.  In the mask perturbation strategy for the visual branch, the masking rate of both textual and visual information is 15\% , while in the adversarial perturbation strategy, the gradient ascent step number is $K=3$, adversarial text and adversarial image perturbation learning rate is $10^{-3}$. Due to limited compute resources, we only conducted the mask perturbation strategy for visual branch contrast learning experiment on the LXMERT model, naming this DCVLP-MASK with LXMERT backbone. However, we applied the mask perturbation strategy for visual branch contrast learning and intra- or inter-contrastive adversarial perturbation strategy for visual branch contrast learning based on the UNITER model, namely, DCVLP-MASK with UNITER-backbone and DCVLP-ADV-intra/inter with UNITER-backbone. Results are given in Tables~\ref{tab:main-1}, \ref{tab:main-2}, and \ref{tab:main-3}.

\subsection{Comparison with State-of-The-Art VLP Methods}
In Table~\ref{tab:main-1}, previous methods adopt masked visual region classification and regression on the visual modality. DCVLP, in contrast, only needs a region contrastive learning loss on the visual modality and shows a competitive effect. DCVLP-MASK with LXMERT backbone and LXMERT have sample training data. DCVLP-MASK with LXMERT backbone outperforms LXMERT by +0.65 on VQA, +0.4 on GQA, and +2.31 on NLVR2. DCVLP-MASK with UNITER-backbone and UNITER-Base have sample training data. Notably, DCVLP-MASK with UNITER-backbone outperforms UNITER-Base by +0.84 on VQA.
VILLA~\cite{gan2020large} is the first to apply large-scale adversarial training for vision-and-language and achieve outstanding performance. VILLA consists of VILLA-Pre and VILLA-Finetune. As in previous research on vision--language pretrained models, we focus on the effects of the pretraining phase. Comparing VILLA-Pre and DCVLP with UNITER-backbone, our model obtains fairly strong results. The core defense strategy of DCVLP-ADV against perturbation is different from VILLA-pre. VILLA-pre utilizes the same loss when generating perturbations and defending against the perturbation. DCVLP-ADV is a new idea for vision--language pretraining, which not only enjoys the advantages of adversarial training but also avoids the use of potentially noisy labels in the adversarial training process. Although the proxy task in DCVLP-ADV is more challenging, the introduction of adversarial training can improve the robustness of the model.

\begin{table}[!htbp]
\centering
\setlength{\abovecaptionskip}{5pt}%
\setlength{\belowcaptionskip}{0pt}%
\scalebox{0.93}{
\begin{tabular}{l|ccc}
\toprule
\multirow{2}{*}{Methods}                                             & \multirow{2}{*}{VQA} & \multirow{2}{*}{RefCOCO+} & \multirow{2}{*}{NLVR2} \\
                                                                    &                      &                      &                        \\ \hline
No Vision Task                                                      & 71.55                & 72.75                & 75.98                 \\ 
Feature Regression                                                  & 71.73                & 73.49                & 76.21                  \\ 
\begin{tabular}[c]{@{}c@{}}Feature Regression \\ + Label\end{tabular} & 71.92                & 74.52                & 76.93                  \\ 
DCVLP-MASK                                                & 72.90                & 75.03                & 77.26                  \\ \bottomrule
\end{tabular}
}
\caption{Comparison of different loss compositions on test-dev.\ splits of VQA, RefCOCO+ and NLVR2 with UNITER in-domain data.}
\vspace*{-3mm}
\label{tab:loss}
\end{table}

\subsection{Ablation Studies and Analyses of DCVLP}
\noindent {\bf Effects of Loss Composition.~}
In Table~\ref{tab:loss}, we perform an ablation study on different loss combinations. 
``No vision task'' method performs visual-linguistic pretraining without adding masks on the visual features. 
After replacing feature regression and label classification loss with contrastive loss, we can achieve the best performance of the ablation study on all three downstream tasks. This consolidates our claim that contrastive learning is more powerful.
Through this implicit method, the visual region can be integrated with richer textual information instead of merely being classified into a limited category.
This is particularly useful when the gap between pretraining and finetuning is large.

\vspace*{2mm}
\noindent {\bf Effects of Different Mask Strategies in UNITER-backbone for DCVLP.~}
From Table~\ref{tab:mask-sys}, we can see that using a symmetric mask strategy yields the best results. Part of the text information in QueryNet is masked, while the complete original text signal is retained in KeyNet. Symmetrically, the complete visual region information is retained in the QueryNet, and part of the visual region information in the KeyNet is masked. We speculate that the advantage of a symmetric mask strategy is to ensure that the visual area can still learn appropriate similar representations under the condition of missing text information and visual information respectively, which promotes the learning ability of both intra-modality and inter-modality. When QueryNet and KeyNet visual modalities use different mask strategies, the model adds too much noise perturbation, which impairs the learning of visual features. Although this strategy can be used to differentiate between positive and negative examples, the region features in this case are not relevant to the ground-truth region features; when the ground-truth feature is present, the feature representation of the positive sample is more similar to the related ground-truth feature, so this situation is more suitable for downstream tasks that do not have the mask region feature.

\vspace*{2mm}
\noindent {\bf Why Cross-Modality Region Contrastive Loss Works.~}
In the pretraining process, QueryNet and KeyNet merge the corresponding text information for each region through an attention mechanism. Under normal circumstances, the corresponding positions of QueryNet and KeyNet for each visual region information should be as similar as possible. However, in our DCVLP model, the input visual information of KeyNet is applied with the mask or adversarial perturbation, so the original visual information is disturbed. In the QueryNet, each visual region feature establishes the intra-modality and inter-modality relationships through self-attention and co-attention; However, because of the mask or adversarial perturbation in the KeyNet, it is more difficult to establish the intra-modality and inter-modality relationships for each visual region feature. Therefore, the use of cross-modality region contrastive loss can encourage QueryNet and KeyNet to generate similar representations. 
\subsection{Visualizing DCVLP Encoder}
In Figure~\ref{vision}, we visualize the attention weights in layers (i.e., the 4th, 8th, 12th layers) of DCVLP with UNITER-backbone. We can see that as the layer grows, the attention weight which indicates correct word-bounding box matching also increases gradually. The visual region not only gradually establishes a relationship with the target text category, but also establishes a certain relationship with related actions, directions, and category words in the text description, which well captures the densely semantic information of the visual region.

\section{Conclusion}
In this paper, we proposed dense contrastive visual-linguistic pretraining (DCVLP) in order to overcome the problems of noisy labels and sparse semantic annotations suffered by existing large-scale multimodal pretraining methods that utilize region feature regression and label classification. Moreover, end-to-end VLP will likely become a mainstream trend for  VLP in the near future, but such an end-to-end training process has no opportunity to draw on classification labels for the image context. Hence, it appears natural and reasonable to combine our DCVLP method into end-to-end VLP training processes. Hence, we also conduct related basic experimental explorations.
\begin{acks}
This work was supported by the Foundation for Innovative Research Groups of the National Natural Science Foundation of China (Grant No. 61921003).
\end{acks}

\bibliographystyle{ACM-Reference-Format}
\bibliography{mm21}


\begin{thebibliography}{54}


\ifx \showCODEN    \undefined \def \showCODEN     #1{\unskip}     \fi
\ifx \showDOI      \undefined \def \showDOI       #1{#1}\fi
\ifx \showISBNx    \undefined \def \showISBNx     #1{\unskip}     \fi
\ifx \showISBNxiii \undefined \def \showISBNxiii  #1{\unskip}     \fi
\ifx \showISSN     \undefined \def \showISSN      #1{\unskip}     \fi
\ifx \showLCCN     \undefined \def \showLCCN      #1{\unskip}     \fi
\ifx \shownote     \undefined \def \shownote      #1{#1}          \fi
\ifx \showarticletitle \undefined \def \showarticletitle #1{#1}   \fi
\ifx \showURL      \undefined \def \showURL       {\relax}        \fi
\providecommand\bibfield[2]{#2}
\providecommand\bibinfo[2]{#2}
\providecommand\natexlab[1]{#1}
\providecommand\showeprint[2][]{arXiv:#2}

\bibitem[\protect\citeauthoryear{Anderson, He, Buehler, Teney, Johnson, Gould,
  and Zhang}{Anderson et~al\mbox{.}}{2018}]%
        {anderson2018bottom}
\bibfield{author}{\bibinfo{person}{Peter Anderson}, \bibinfo{person}{Xiaodong
  He}, \bibinfo{person}{Chris Buehler}, \bibinfo{person}{Damien Teney},
  \bibinfo{person}{Mark Johnson}, \bibinfo{person}{Stephen Gould}, {and}
  \bibinfo{person}{Lei Zhang}.} \bibinfo{year}{2018}\natexlab{}.
\newblock \showarticletitle{Bottom-up and top-down attention for image
  captioning and visual question answering}. In
  \bibinfo{booktitle}{\emph{Proceedings of the IEEE conference on computer
  vision and pattern recognition}}. \bibinfo{pages}{6077--6086}.
\newblock


\bibitem[\protect\citeauthoryear{Bulent~Sariyildiz, Perez, and
  Larlus}{Bulent~Sariyildiz et~al\mbox{.}}{2020}]%
        {bulent2020learning}
\bibfield{author}{\bibinfo{person}{Mert Bulent~Sariyildiz},
  \bibinfo{person}{Julien Perez}, {and} \bibinfo{person}{Diane Larlus}.}
  \bibinfo{year}{2020}\natexlab{}.
\newblock \showarticletitle{Learning Visual Representations with Caption
  Annotations}.
\newblock \bibinfo{journal}{\emph{arXiv e-prints}} (\bibinfo{year}{2020}),
  \bibinfo{pages}{arXiv--2008}.
\newblock


\bibitem[\protect\citeauthoryear{Chen, Kornblith, Norouzi, and Hinton}{Chen
  et~al\mbox{.}}{2020b}]%
        {chen2020simple}
\bibfield{author}{\bibinfo{person}{Ting Chen}, \bibinfo{person}{Simon
  Kornblith}, \bibinfo{person}{Mohammad Norouzi}, {and}
  \bibinfo{person}{Geoffrey Hinton}.} \bibinfo{year}{2020}\natexlab{b}.
\newblock \showarticletitle{A simple framework for contrastive learning of
  visual representations}.
\newblock \bibinfo{journal}{\emph{arXiv preprint arXiv:2002.05709}}
  (\bibinfo{year}{2020}).
\newblock


\bibitem[\protect\citeauthoryear{Chen, Fan, Girshick, and He}{Chen
  et~al\mbox{.}}{2020a}]%
        {chen2020improved}
\bibfield{author}{\bibinfo{person}{Xinlei Chen}, \bibinfo{person}{Haoqi Fan},
  \bibinfo{person}{Ross Girshick}, {and} \bibinfo{person}{Kaiming He}.}
  \bibinfo{year}{2020}\natexlab{a}.
\newblock \showarticletitle{Improved Baselines with Momentum Contrastive
  Learning}.
\newblock \bibinfo{journal}{\emph{arXiv preprint arXiv:2003.04297}}
  (\bibinfo{year}{2020}).
\newblock


\bibitem[\protect\citeauthoryear{Chen, Li, Yu, Kholy, Ahmed, Gan, Cheng, and
  Liu}{Chen et~al\mbox{.}}{2019}]%
        {chen2019uniter}
\bibfield{author}{\bibinfo{person}{Yen-Chun Chen}, \bibinfo{person}{Linjie Li},
  \bibinfo{person}{Licheng Yu}, \bibinfo{person}{Ahmed~El Kholy},
  \bibinfo{person}{Faisal Ahmed}, \bibinfo{person}{Zhe Gan},
  \bibinfo{person}{Yu Cheng}, {and} \bibinfo{person}{Jingjing Liu}.}
  \bibinfo{year}{2019}\natexlab{}.
\newblock \showarticletitle{Uniter: Learning universal image-text
  representations}.
\newblock \bibinfo{journal}{\emph{arXiv preprint arXiv:1909.11740}}
  (\bibinfo{year}{2019}).
\newblock


\bibitem[\protect\citeauthoryear{Deng, Dong, Socher, Li, Li, and Fei-Fei}{Deng
  et~al\mbox{.}}{2009}]%
        {deng2009imagenet}
\bibfield{author}{\bibinfo{person}{Jia Deng}, \bibinfo{person}{Wei Dong},
  \bibinfo{person}{Richard Socher}, \bibinfo{person}{Li-Jia Li},
  \bibinfo{person}{Kai Li}, {and} \bibinfo{person}{Li Fei-Fei}.}
  \bibinfo{year}{2009}\natexlab{}.
\newblock \showarticletitle{Imagenet: A large-scale hierarchical image
  database}. In \bibinfo{booktitle}{\emph{2009 IEEE conference on computer
  vision and pattern recognition}}. Ieee, \bibinfo{pages}{248--255}.
\newblock


\bibitem[\protect\citeauthoryear{Desai and Johnson}{Desai and Johnson}{2020}]%
        {desai2020virtex}
\bibfield{author}{\bibinfo{person}{Karan Desai} {and} \bibinfo{person}{Justin
  Johnson}.} \bibinfo{year}{2020}\natexlab{}.
\newblock \showarticletitle{VirTex: Learning Visual Representations from
  Textual Annotations}.
\newblock \bibinfo{journal}{\emph{arXiv preprint arXiv:2006.06666}}
  (\bibinfo{year}{2020}).
\newblock


\bibitem[\protect\citeauthoryear{Devlin, Chang, Lee, and Toutanova}{Devlin
  et~al\mbox{.}}{2018}]%
        {devlin2018bert}
\bibfield{author}{\bibinfo{person}{Jacob Devlin}, \bibinfo{person}{Ming-Wei
  Chang}, \bibinfo{person}{Kenton Lee}, {and} \bibinfo{person}{Kristina
  Toutanova}.} \bibinfo{year}{2018}\natexlab{}.
\newblock \showarticletitle{Bert: Pre-training of deep bidirectional
  transformers for language understanding}.
\newblock \bibinfo{journal}{\emph{arXiv preprint arXiv:1810.04805}}
  (\bibinfo{year}{2018}).
\newblock


\bibitem[\protect\citeauthoryear{Gan, Chen, Li, Zhu, Cheng, and Liu}{Gan
  et~al\mbox{.}}{2020}]%
        {gan2020large}
\bibfield{author}{\bibinfo{person}{Zhe Gan}, \bibinfo{person}{Yen-Chun Chen},
  \bibinfo{person}{Linjie Li}, \bibinfo{person}{Chen Zhu}, \bibinfo{person}{Yu
  Cheng}, {and} \bibinfo{person}{Jingjing Liu}.}
  \bibinfo{year}{2020}\natexlab{}.
\newblock \showarticletitle{Large-Scale Adversarial Training for
  Vision-and-Language Representation Learning}.
\newblock \bibinfo{journal}{\emph{arXiv preprint arXiv:2006.06195}}
  (\bibinfo{year}{2020}).
\newblock


\bibitem[\protect\citeauthoryear{Gao, Jiang, You, Lu, Hoi, Wang, and Li}{Gao
  et~al\mbox{.}}{2019}]%
        {gao2019dynamic}
\bibfield{author}{\bibinfo{person}{Peng Gao}, \bibinfo{person}{Zhengkai Jiang},
  \bibinfo{person}{Haoxuan You}, \bibinfo{person}{Pan Lu},
  \bibinfo{person}{Steven~CH Hoi}, \bibinfo{person}{Xiaogang Wang}, {and}
  \bibinfo{person}{Hongsheng Li}.} \bibinfo{year}{2019}\natexlab{}.
\newblock \showarticletitle{Dynamic fusion with intra-and inter-modality
  attention flow for visual question answering}. In
  \bibinfo{booktitle}{\emph{Proceedings of the IEEE Conference on Computer
  Vision and Pattern Recognition}}. \bibinfo{pages}{6639--6648}.
\newblock


\bibitem[\protect\citeauthoryear{Geng, Gao, Chatterjee, Hori, Le~Roux, Zhang,
  Li, and Cherian}{Geng et~al\mbox{.}}{2021}]%
        {geng2021dynamic}
\bibfield{author}{\bibinfo{person}{Shijie Geng}, \bibinfo{person}{Peng Gao},
  \bibinfo{person}{Moitreya Chatterjee}, \bibinfo{person}{Chiori Hori},
  \bibinfo{person}{Jonathan Le~Roux}, \bibinfo{person}{Yongfeng Zhang},
  \bibinfo{person}{Hongsheng Li}, {and} \bibinfo{person}{Anoop Cherian}.}
  \bibinfo{year}{2021}\natexlab{}.
\newblock \showarticletitle{Dynamic graph representation learning for video
  dialog via multi-modal shuffled transformers}. In
  \bibinfo{booktitle}{\emph{AAAI Conference on Artificial Intelligence}}.
\newblock


\bibitem[\protect\citeauthoryear{Goodfellow, Shlens, and Szegedy}{Goodfellow
  et~al\mbox{.}}{2014}]%
        {goodfellow2014explaining}
\bibfield{author}{\bibinfo{person}{Ian~J Goodfellow}, \bibinfo{person}{Jonathon
  Shlens}, {and} \bibinfo{person}{Christian Szegedy}.}
  \bibinfo{year}{2014}\natexlab{}.
\newblock \showarticletitle{Explaining and harnessing adversarial examples}.
\newblock \bibinfo{journal}{\emph{arXiv preprint arXiv:1412.6572}}
  (\bibinfo{year}{2014}).
\newblock


\bibitem[\protect\citeauthoryear{Goyal, Khot, Summers{-}Stay, Batra, and
  Parikh}{Goyal et~al\mbox{.}}{2017}]%
        {balanced_vqa_v2}
\bibfield{author}{\bibinfo{person}{Yash Goyal}, \bibinfo{person}{Tejas Khot},
  \bibinfo{person}{Douglas Summers{-}Stay}, \bibinfo{person}{Dhruv Batra},
  {and} \bibinfo{person}{Devi Parikh}.} \bibinfo{year}{2017}\natexlab{}.
\newblock \showarticletitle{Making the {V} in {VQA} Matter: Elevating the Role
  of Image Understanding in {V}isual {Q}uestion {A}nswering}. In
  \bibinfo{booktitle}{\emph{Conference on Computer Vision and Pattern
  Recognition (CVPR)}}.
\newblock


\bibitem[\protect\citeauthoryear{Grill, Strub, Altch{\'e}, Tallec, Richemond,
  Buchatskaya, Doersch, Pires, Guo, Azar, et~al\mbox{.}}{Grill
  et~al\mbox{.}}{2020}]%
        {grill2020bootstrap}
\bibfield{author}{\bibinfo{person}{Jean-Bastien Grill},
  \bibinfo{person}{Florian Strub}, \bibinfo{person}{Florent Altch{\'e}},
  \bibinfo{person}{Corentin Tallec}, \bibinfo{person}{Pierre~H Richemond},
  \bibinfo{person}{Elena Buchatskaya}, \bibinfo{person}{Carl Doersch},
  \bibinfo{person}{Bernardo~Avila Pires}, \bibinfo{person}{Zhaohan~Daniel Guo},
  \bibinfo{person}{Mohammad~Gheshlaghi Azar}, {et~al\mbox{.}}}
  \bibinfo{year}{2020}\natexlab{}.
\newblock \showarticletitle{Bootstrap your own latent: A new approach to
  self-supervised learning}.
\newblock \bibinfo{journal}{\emph{arXiv preprint arXiv:2006.07733}}
  (\bibinfo{year}{2020}).
\newblock


\bibitem[\protect\citeauthoryear{Hadsell, Chopra, and LeCun}{Hadsell
  et~al\mbox{.}}{2006}]%
        {hadsell2006dimensionality}
\bibfield{author}{\bibinfo{person}{Raia Hadsell}, \bibinfo{person}{Sumit
  Chopra}, {and} \bibinfo{person}{Yann LeCun}.}
  \bibinfo{year}{2006}\natexlab{}.
\newblock \showarticletitle{Dimensionality reduction by learning an invariant
  mapping}. In \bibinfo{booktitle}{\emph{2006 IEEE Computer Society Conference
  on Computer Vision and Pattern Recognition (CVPR'06)}},
  Vol.~\bibinfo{volume}{2}. IEEE, \bibinfo{pages}{1735--1742}.
\newblock


\bibitem[\protect\citeauthoryear{He, Fan, Wu, Xie, and Girshick}{He
  et~al\mbox{.}}{2019}]%
        {he2019momentum}
\bibfield{author}{\bibinfo{person}{Kaiming He}, \bibinfo{person}{Haoqi Fan},
  \bibinfo{person}{Yuxin Wu}, \bibinfo{person}{Saining Xie}, {and}
  \bibinfo{person}{Ross Girshick}.} \bibinfo{year}{2019}\natexlab{}.
\newblock \showarticletitle{Momentum contrast for unsupervised visual
  representation learning}.
\newblock \bibinfo{journal}{\emph{arXiv preprint arXiv:1911.05722}}
  (\bibinfo{year}{2019}).
\newblock


\bibitem[\protect\citeauthoryear{Hu, Wang, Hu, and Qi}{Hu
  et~al\mbox{.}}{2020}]%
        {hu2020adco}
\bibfield{author}{\bibinfo{person}{Qianjiang Hu}, \bibinfo{person}{Xiao Wang},
  \bibinfo{person}{Wei Hu}, {and} \bibinfo{person}{Guo-Jun Qi}.}
  \bibinfo{year}{2020}\natexlab{}.
\newblock \showarticletitle{AdCo: Adversarial Contrast for Efficient Learning
  of Unsupervised Representations from Self-Trained Negative Adversaries}.
\newblock \bibinfo{journal}{\emph{arXiv preprint arXiv:2011.08435}}
  (\bibinfo{year}{2020}).
\newblock


\bibitem[\protect\citeauthoryear{Hudson and Manning}{Hudson and
  Manning}{2019}]%
        {hudson2019gqa}
\bibfield{author}{\bibinfo{person}{Drew~A Hudson} {and}
  \bibinfo{person}{Christopher~D Manning}.} \bibinfo{year}{2019}\natexlab{}.
\newblock \showarticletitle{GQA: A New Dataset for Real-World Visual Reasoning
  and Compositional Question Answering}.
\newblock \bibinfo{journal}{\emph{Conference on Computer Vision and Pattern
  Recognition (CVPR)}} (\bibinfo{year}{2019}).
\newblock


\bibitem[\protect\citeauthoryear{Jiang, Chen, Chen, and Wang}{Jiang
  et~al\mbox{.}}{2020}]%
        {jiang2020robust}
\bibfield{author}{\bibinfo{person}{Ziyu Jiang}, \bibinfo{person}{Tianlong
  Chen}, \bibinfo{person}{Ting Chen}, {and} \bibinfo{person}{Zhangyang Wang}.}
  \bibinfo{year}{2020}\natexlab{}.
\newblock \showarticletitle{Robust pre-training by adversarial contrastive
  learning}.
\newblock \bibinfo{journal}{\emph{arXiv preprint arXiv:2010.13337}}
  (\bibinfo{year}{2020}).
\newblock


\bibitem[\protect\citeauthoryear{Kolesnikov, Zhai, and Beyer}{Kolesnikov
  et~al\mbox{.}}{2019}]%
        {kolesnikov2019revisiting}
\bibfield{author}{\bibinfo{person}{Alexander Kolesnikov},
  \bibinfo{person}{Xiaohua Zhai}, {and} \bibinfo{person}{Lucas Beyer}.}
  \bibinfo{year}{2019}\natexlab{}.
\newblock \showarticletitle{Revisiting self-supervised visual representation
  learning}. In \bibinfo{booktitle}{\emph{Proceedings of the IEEE conference on
  Computer Vision and Pattern Recognition}}. \bibinfo{pages}{1920--1929}.
\newblock


\bibitem[\protect\citeauthoryear{Krishna, Zhu, Groth, Johnson, Hata, Kravitz,
  Chen, Kalantidis, Li, Shamma, et~al\mbox{.}}{Krishna et~al\mbox{.}}{2017}]%
        {krishna2017visual}
\bibfield{author}{\bibinfo{person}{Ranjay Krishna}, \bibinfo{person}{Yuke Zhu},
  \bibinfo{person}{Oliver Groth}, \bibinfo{person}{Justin Johnson},
  \bibinfo{person}{Kenji Hata}, \bibinfo{person}{Joshua Kravitz},
  \bibinfo{person}{Stephanie Chen}, \bibinfo{person}{Yannis Kalantidis},
  \bibinfo{person}{Li-Jia Li}, \bibinfo{person}{David~A Shamma},
  {et~al\mbox{.}}} \bibinfo{year}{2017}\natexlab{}.
\newblock \showarticletitle{Visual genome: Connecting language and vision using
  crowdsourced dense image annotations}.
\newblock \bibinfo{journal}{\emph{International Journal of Computer Vision}}
  \bibinfo{volume}{123}, \bibinfo{number}{1} (\bibinfo{year}{2017}),
  \bibinfo{pages}{32--73}.
\newblock


\bibitem[\protect\citeauthoryear{Krizhevsky, Sutskever, and Hinton}{Krizhevsky
  et~al\mbox{.}}{2012}]%
        {krizhevsky2012imagenet}
\bibfield{author}{\bibinfo{person}{Alex Krizhevsky}, \bibinfo{person}{Ilya
  Sutskever}, {and} \bibinfo{person}{Geoffrey~E Hinton}.}
  \bibinfo{year}{2012}\natexlab{}.
\newblock \showarticletitle{Imagenet classification with deep convolutional
  neural networks}. In \bibinfo{booktitle}{\emph{Advances in neural information
  processing systems}}. \bibinfo{pages}{1097--1105}.
\newblock


\bibitem[\protect\citeauthoryear{Lee, Chen, Hua, Hu, and He}{Lee
  et~al\mbox{.}}{2018}]%
        {lee2018stacked}
\bibfield{author}{\bibinfo{person}{Kuang-Huei Lee}, \bibinfo{person}{Xi Chen},
  \bibinfo{person}{Gang Hua}, \bibinfo{person}{Houdong Hu}, {and}
  \bibinfo{person}{Xiaodong He}.} \bibinfo{year}{2018}\natexlab{}.
\newblock \showarticletitle{Stacked cross attention for image-text matching}.
  In \bibinfo{booktitle}{\emph{Proceedings of the European Conference on
  Computer Vision (ECCV)}}. \bibinfo{pages}{201--216}.
\newblock


\bibitem[\protect\citeauthoryear{Li, Gan, Cheng, and Liu}{Li
  et~al\mbox{.}}{2019a}]%
        {li2019relation}
\bibfield{author}{\bibinfo{person}{Linjie Li}, \bibinfo{person}{Zhe Gan},
  \bibinfo{person}{Yu Cheng}, {and} \bibinfo{person}{Jingjing Liu}.}
  \bibinfo{year}{2019}\natexlab{a}.
\newblock \showarticletitle{Relation-aware graph attention network for visual
  question answering}. In \bibinfo{booktitle}{\emph{Proceedings of the IEEE
  International Conference on Computer Vision}}. \bibinfo{pages}{10313--10322}.
\newblock


\bibitem[\protect\citeauthoryear{Li, Yatskar, Yin, Hsieh, and Chang}{Li
  et~al\mbox{.}}{2019b}]%
        {li2019visualbert}
\bibfield{author}{\bibinfo{person}{Liunian~Harold Li}, \bibinfo{person}{Mark
  Yatskar}, \bibinfo{person}{Da Yin}, \bibinfo{person}{Cho-Jui Hsieh}, {and}
  \bibinfo{person}{Kai-Wei Chang}.} \bibinfo{year}{2019}\natexlab{b}.
\newblock \showarticletitle{Visualbert: A simple and performant baseline for
  vision and language}.
\newblock \bibinfo{journal}{\emph{arXiv preprint arXiv:1908.03557}}
  (\bibinfo{year}{2019}).
\newblock


\bibitem[\protect\citeauthoryear{Li, Yin, Li, Hu, Zhang, Zhang, Wang, Hu, Dong,
  Wei, et~al\mbox{.}}{Li et~al\mbox{.}}{2020}]%
        {li2020oscar}
\bibfield{author}{\bibinfo{person}{Xiujun Li}, \bibinfo{person}{Xi Yin},
  \bibinfo{person}{Chunyuan Li}, \bibinfo{person}{Xiaowei Hu},
  \bibinfo{person}{Pengchuan Zhang}, \bibinfo{person}{Lei Zhang},
  \bibinfo{person}{Lijuan Wang}, \bibinfo{person}{Houdong Hu},
  \bibinfo{person}{Li Dong}, \bibinfo{person}{Furu Wei}, {et~al\mbox{.}}}
  \bibinfo{year}{2020}\natexlab{}.
\newblock \showarticletitle{Oscar: Object-semantics aligned pre-training for
  vision-language tasks}.
\newblock \bibinfo{journal}{\emph{arXiv preprint arXiv:2004.06165}}
  (\bibinfo{year}{2020}).
\newblock


\bibitem[\protect\citeauthoryear{Lin, Maire, Belongie, Hays, Perona, Ramanan,
  Doll{\'a}r, and Zitnick}{Lin et~al\mbox{.}}{2014}]%
        {lin2014microsoft}
\bibfield{author}{\bibinfo{person}{Tsung-Yi Lin}, \bibinfo{person}{Michael
  Maire}, \bibinfo{person}{Serge Belongie}, \bibinfo{person}{James Hays},
  \bibinfo{person}{Pietro Perona}, \bibinfo{person}{Deva Ramanan},
  \bibinfo{person}{Piotr Doll{\'a}r}, {and} \bibinfo{person}{C~Lawrence
  Zitnick}.} \bibinfo{year}{2014}\natexlab{}.
\newblock \showarticletitle{Microsoft coco: Common objects in context}. In
  \bibinfo{booktitle}{\emph{European conference on computer vision}}. Springer,
  \bibinfo{pages}{740--755}.
\newblock


\bibitem[\protect\citeauthoryear{Liu, Ott, Goyal, Du, Joshi, Chen, Levy, Lewis,
  Zettlemoyer, and Stoyanov}{Liu et~al\mbox{.}}{2019}]%
        {liu2019roberta}
\bibfield{author}{\bibinfo{person}{Yinhan Liu}, \bibinfo{person}{Myle Ott},
  \bibinfo{person}{Naman Goyal}, \bibinfo{person}{Jingfei Du},
  \bibinfo{person}{Mandar Joshi}, \bibinfo{person}{Danqi Chen},
  \bibinfo{person}{Omer Levy}, \bibinfo{person}{Mike Lewis},
  \bibinfo{person}{Luke Zettlemoyer}, {and} \bibinfo{person}{Veselin
  Stoyanov}.} \bibinfo{year}{2019}\natexlab{}.
\newblock \showarticletitle{Roberta: A robustly optimized bert pretraining
  approach}.
\newblock \bibinfo{journal}{\emph{arXiv preprint arXiv:1907.11692}}
  (\bibinfo{year}{2019}).
\newblock


\bibitem[\protect\citeauthoryear{Lu, Batra, Parikh, and Lee}{Lu
  et~al\mbox{.}}{2019}]%
        {lu2019vilbert}
\bibfield{author}{\bibinfo{person}{Jiasen Lu}, \bibinfo{person}{Dhruv Batra},
  \bibinfo{person}{Devi Parikh}, {and} \bibinfo{person}{Stefan Lee}.}
  \bibinfo{year}{2019}\natexlab{}.
\newblock \showarticletitle{Vilbert: Pretraining task-agnostic visiolinguistic
  representations for vision-and-language tasks}. In
  \bibinfo{booktitle}{\emph{Advances in Neural Information Processing
  Systems}}. \bibinfo{pages}{13--23}.
\newblock


\bibitem[\protect\citeauthoryear{Madry, Makelov, Schmidt, Tsipras, and
  Vladu}{Madry et~al\mbox{.}}{2018}]%
        {madry2018towards}
\bibfield{author}{\bibinfo{person}{Aleksander Madry},
  \bibinfo{person}{Aleksandar Makelov}, \bibinfo{person}{Ludwig Schmidt},
  \bibinfo{person}{Dimitris Tsipras}, {and} \bibinfo{person}{Adrian Vladu}.}
  \bibinfo{year}{2018}\natexlab{}.
\newblock \showarticletitle{Towards Deep Learning Models Resistant to
  Adversarial Attacks}. In \bibinfo{booktitle}{\emph{International Conference
  on Learning Representations}}.
\newblock
\urldef\tempurl%
\url{https://openreview.net/forum?id=rJzIBfZAb}
\showURL{%
\tempurl}


\bibitem[\protect\citeauthoryear{Moosavi-Dezfooli, Fawzi, Fawzi, and
  Frossard}{Moosavi-Dezfooli et~al\mbox{.}}{2017}]%
        {moosavi2017universal}
\bibfield{author}{\bibinfo{person}{Seyed-Mohsen Moosavi-Dezfooli},
  \bibinfo{person}{Alhussein Fawzi}, \bibinfo{person}{Omar Fawzi}, {and}
  \bibinfo{person}{Pascal Frossard}.} \bibinfo{year}{2017}\natexlab{}.
\newblock \showarticletitle{Universal adversarial perturbations}. In
  \bibinfo{booktitle}{\emph{Proceedings of the IEEE conference on computer
  vision and pattern recognition}}. \bibinfo{pages}{1765--1773}.
\newblock


\bibitem[\protect\citeauthoryear{Nagaraja, Morariu, and Davis}{Nagaraja
  et~al\mbox{.}}{2016}]%
        {nagaraja2016modeling}
\bibfield{author}{\bibinfo{person}{Varun~K Nagaraja}, \bibinfo{person}{Vlad~I
  Morariu}, {and} \bibinfo{person}{Larry~S Davis}.}
  \bibinfo{year}{2016}\natexlab{}.
\newblock \showarticletitle{Modeling context between objects for referring
  expression understanding}. In \bibinfo{booktitle}{\emph{European Conference
  on Computer Vision}}. Springer, \bibinfo{pages}{792--807}.
\newblock


\bibitem[\protect\citeauthoryear{Noroozi and Favaro}{Noroozi and
  Favaro}{2016}]%
        {noroozi2016unsupervised}
\bibfield{author}{\bibinfo{person}{Mehdi Noroozi} {and} \bibinfo{person}{Paolo
  Favaro}.} \bibinfo{year}{2016}\natexlab{}.
\newblock \showarticletitle{Unsupervised learning of visual representations by
  solving jigsaw puzzles}. In \bibinfo{booktitle}{\emph{European Conference on
  Computer Vision}}. Springer, \bibinfo{pages}{69--84}.
\newblock


\bibitem[\protect\citeauthoryear{Ordonez, Kulkarni, and Berg}{Ordonez
  et~al\mbox{.}}{2011}]%
        {ordonez2011im2text}
\bibfield{author}{\bibinfo{person}{Vicente Ordonez}, \bibinfo{person}{Girish
  Kulkarni}, {and} \bibinfo{person}{Tamara Berg}.}
  \bibinfo{year}{2011}\natexlab{}.
\newblock \showarticletitle{Im2text: Describing images using 1 million
  captioned photographs}.
\newblock \bibinfo{journal}{\emph{Advances in neural information processing
  systems}}  \bibinfo{volume}{24} (\bibinfo{year}{2011}),
  \bibinfo{pages}{1143--1151}.
\newblock


\bibitem[\protect\citeauthoryear{Pathak, Krahenbuhl, Donahue, Darrell, and
  Efros}{Pathak et~al\mbox{.}}{2016}]%
        {pathak2016context}
\bibfield{author}{\bibinfo{person}{Deepak Pathak}, \bibinfo{person}{Philipp
  Krahenbuhl}, \bibinfo{person}{Jeff Donahue}, \bibinfo{person}{Trevor
  Darrell}, {and} \bibinfo{person}{Alexei~A Efros}.}
  \bibinfo{year}{2016}\natexlab{}.
\newblock \showarticletitle{Context encoders: Feature learning by inpainting}.
  In \bibinfo{booktitle}{\emph{Proceedings of the IEEE conference on computer
  vision and pattern recognition}}. \bibinfo{pages}{2536--2544}.
\newblock


\bibitem[\protect\citeauthoryear{Radford, Kim, Hallacy, Ramesh, Goh, Agarwal,
  Sastry, Askell, Mishkin, Clark, et~al\mbox{.}}{Radford et~al\mbox{.}}{2021}]%
        {radford2021learning}
\bibfield{author}{\bibinfo{person}{Alec Radford}, \bibinfo{person}{Jong~Wook
  Kim}, \bibinfo{person}{Chris Hallacy}, \bibinfo{person}{Aditya Ramesh},
  \bibinfo{person}{Gabriel Goh}, \bibinfo{person}{Sandhini Agarwal},
  \bibinfo{person}{Girish Sastry}, \bibinfo{person}{Amanda Askell},
  \bibinfo{person}{Pamela Mishkin}, \bibinfo{person}{Jack Clark},
  {et~al\mbox{.}}} \bibinfo{year}{2021}\natexlab{}.
\newblock \showarticletitle{Learning transferable visual models from natural
  language supervision}.
\newblock \bibinfo{journal}{\emph{arXiv preprint arXiv:2103.00020}}
  (\bibinfo{year}{2021}).
\newblock


\bibitem[\protect\citeauthoryear{Radford, Narasimhan, Salimans, and
  Sutskever}{Radford et~al\mbox{.}}{2018}]%
        {radford2018improving}
\bibfield{author}{\bibinfo{person}{Alec Radford}, \bibinfo{person}{Karthik
  Narasimhan}, \bibinfo{person}{Tim Salimans}, {and} \bibinfo{person}{Ilya
  Sutskever}.} \bibinfo{year}{2018}\natexlab{}.
\newblock \showarticletitle{Improving language understanding by generative
  pre-training}.
\newblock  (\bibinfo{year}{2018}).
\newblock


\bibitem[\protect\citeauthoryear{Ren, Ren, Liao, Liu, Li, Han, and Yan}{Ren
  et~al\mbox{.}}{2020}]%
        {ren2020scene}
\bibfield{author}{\bibinfo{person}{Guanghui Ren}, \bibinfo{person}{Lejian Ren},
  \bibinfo{person}{Yue Liao}, \bibinfo{person}{Si Liu}, \bibinfo{person}{Bo
  Li}, \bibinfo{person}{Jizhong Han}, {and} \bibinfo{person}{Shuicheng Yan}.}
  \bibinfo{year}{2020}\natexlab{}.
\newblock \showarticletitle{Scene graph generation with hierarchical context}.
\newblock \bibinfo{journal}{\emph{IEEE Transactions on Neural Networks and
  Learning Systems}} \bibinfo{volume}{32}, \bibinfo{number}{2}
  (\bibinfo{year}{2020}), \bibinfo{pages}{909--915}.
\newblock


\bibitem[\protect\citeauthoryear{Ren, He, Girshick, and Sun}{Ren
  et~al\mbox{.}}{2015}]%
        {ren2015faster}
\bibfield{author}{\bibinfo{person}{Shaoqing Ren}, \bibinfo{person}{Kaiming He},
  \bibinfo{person}{Ross Girshick}, {and} \bibinfo{person}{Jian Sun}.}
  \bibinfo{year}{2015}\natexlab{}.
\newblock \showarticletitle{Faster r-cnn: Towards real-time object detection
  with region proposal networks}. In \bibinfo{booktitle}{\emph{Advances in
  neural information processing systems}}. \bibinfo{pages}{91--99}.
\newblock


\bibitem[\protect\citeauthoryear{Shafahi, Najibi, Ghiasi, Xu, Dickerson,
  Studer, Davis, Taylor, and Goldstein}{Shafahi et~al\mbox{.}}{2019}]%
        {shafahi2019adversarial}
\bibfield{author}{\bibinfo{person}{Ali Shafahi}, \bibinfo{person}{Mahyar
  Najibi}, \bibinfo{person}{Amin Ghiasi}, \bibinfo{person}{Zheng Xu},
  \bibinfo{person}{John Dickerson}, \bibinfo{person}{Christoph Studer},
  \bibinfo{person}{Larry~S Davis}, \bibinfo{person}{Gavin Taylor}, {and}
  \bibinfo{person}{Tom Goldstein}.} \bibinfo{year}{2019}\natexlab{}.
\newblock \showarticletitle{Adversarial training for free!}
\newblock \bibinfo{journal}{\emph{arXiv preprint arXiv:1904.12843}}
  (\bibinfo{year}{2019}).
\newblock


\bibitem[\protect\citeauthoryear{Sharma, Ding, Goodman, and Soricut}{Sharma
  et~al\mbox{.}}{2018}]%
        {sharma2018conceptual}
\bibfield{author}{\bibinfo{person}{Piyush Sharma}, \bibinfo{person}{Nan Ding},
  \bibinfo{person}{Sebastian Goodman}, {and} \bibinfo{person}{Radu Soricut}.}
  \bibinfo{year}{2018}\natexlab{}.
\newblock \showarticletitle{Conceptual captions: A cleaned, hypernymed, image
  alt-text dataset for automatic image captioning}. In
  \bibinfo{booktitle}{\emph{Proceedings of the 56th Annual Meeting of the
  Association for Computational Linguistics (Volume 1: Long Papers)}}.
  \bibinfo{pages}{2556--2565}.
\newblock


\bibitem[\protect\citeauthoryear{Shi, Geng, Shuang, Hori, Liu, Gao, and Su}{Shi
  et~al\mbox{.}}{2020}]%
        {shi2020multi}
\bibfield{author}{\bibinfo{person}{Lei Shi}, \bibinfo{person}{Shijie Geng},
  \bibinfo{person}{Kai Shuang}, \bibinfo{person}{Chiori Hori},
  \bibinfo{person}{Songxiang Liu}, \bibinfo{person}{Peng Gao}, {and}
  \bibinfo{person}{Sen Su}.} \bibinfo{year}{2020}\natexlab{}.
\newblock \showarticletitle{Multi-Layer Content Interaction Through Quaternion
  Product For Visual Question Answering}. In \bibinfo{booktitle}{\emph{IEEE
  International Conference on Acoustics, Speech and Signal Processing
  (ICASSP)}}. IEEE, \bibinfo{pages}{4412--4416}.
\newblock


\bibitem[\protect\citeauthoryear{Suhr, Lewis, Yeh, and Artzi}{Suhr
  et~al\mbox{.}}{2017}]%
        {suhr2017corpus}
\bibfield{author}{\bibinfo{person}{Alane Suhr}, \bibinfo{person}{Mike Lewis},
  \bibinfo{person}{James Yeh}, {and} \bibinfo{person}{Yoav Artzi}.}
  \bibinfo{year}{2017}\natexlab{}.
\newblock \showarticletitle{A corpus of natural language for visual reasoning}.
  In \bibinfo{booktitle}{\emph{Proceedings of the 55th Annual Meeting of the
  Association for Computational Linguistics (Volume 2: Short Papers)}}.
  \bibinfo{pages}{217--223}.
\newblock


\bibitem[\protect\citeauthoryear{Suhr, Zhou, Zhang, Zhang, Bai, and Artzi}{Suhr
  et~al\mbox{.}}{2018}]%
        {suhr2018corpus}
\bibfield{author}{\bibinfo{person}{Alane Suhr}, \bibinfo{person}{Stephanie
  Zhou}, \bibinfo{person}{Ally Zhang}, \bibinfo{person}{Iris Zhang},
  \bibinfo{person}{Huajun Bai}, {and} \bibinfo{person}{Yoav Artzi}.}
  \bibinfo{year}{2018}\natexlab{}.
\newblock \showarticletitle{A corpus for reasoning about natural language
  grounded in photographs}.
\newblock \bibinfo{journal}{\emph{arXiv preprint arXiv:1811.00491}}
  (\bibinfo{year}{2018}).
\newblock


\bibitem[\protect\citeauthoryear{Szegedy, Zaremba, Sutskever, Bruna, Erhan,
  Goodfellow, and Fergus}{Szegedy et~al\mbox{.}}{2013}]%
        {szegedy2013intriguing}
\bibfield{author}{\bibinfo{person}{Christian Szegedy},
  \bibinfo{person}{Wojciech Zaremba}, \bibinfo{person}{Ilya Sutskever},
  \bibinfo{person}{Joan Bruna}, \bibinfo{person}{Dumitru Erhan},
  \bibinfo{person}{Ian Goodfellow}, {and} \bibinfo{person}{Rob Fergus}.}
  \bibinfo{year}{2013}\natexlab{}.
\newblock \showarticletitle{Intriguing properties of neural networks}.
\newblock \bibinfo{journal}{\emph{arXiv preprint arXiv:1312.6199}}
  (\bibinfo{year}{2013}).
\newblock


\bibitem[\protect\citeauthoryear{Tan and Bansal}{Tan and Bansal}{2019}]%
        {tan2019lxmert}
\bibfield{author}{\bibinfo{person}{Hao Tan} {and} \bibinfo{person}{Mohit
  Bansal}.} \bibinfo{year}{2019}\natexlab{}.
\newblock \showarticletitle{Lxmert: Learning cross-modality encoder
  representations from transformers}.
\newblock \bibinfo{journal}{\emph{arXiv preprint arXiv:1908.07490}}
  (\bibinfo{year}{2019}).
\newblock


\bibitem[\protect\citeauthoryear{Tian, Krishnan, and Isola}{Tian
  et~al\mbox{.}}{2019}]%
        {tian2019contrastive}
\bibfield{author}{\bibinfo{person}{Yonglong Tian}, \bibinfo{person}{Dilip
  Krishnan}, {and} \bibinfo{person}{Phillip Isola}.}
  \bibinfo{year}{2019}\natexlab{}.
\newblock \showarticletitle{Contrastive multiview coding}.
\newblock \bibinfo{journal}{\emph{arXiv preprint arXiv:1906.05849}}
  (\bibinfo{year}{2019}).
\newblock


\bibitem[\protect\citeauthoryear{Vaswani, Shazeer, Parmar, Uszkoreit, Jones,
  Gomez, Kaiser, and Polosukhin}{Vaswani et~al\mbox{.}}{2017}]%
        {vaswani2017attention}
\bibfield{author}{\bibinfo{person}{Ashish Vaswani}, \bibinfo{person}{Noam
  Shazeer}, \bibinfo{person}{Niki Parmar}, \bibinfo{person}{Jakob Uszkoreit},
  \bibinfo{person}{Llion Jones}, \bibinfo{person}{Aidan~N Gomez},
  \bibinfo{person}{{\L}ukasz Kaiser}, {and} \bibinfo{person}{Illia
  Polosukhin}.} \bibinfo{year}{2017}\natexlab{}.
\newblock \showarticletitle{Attention is all you need}. In
  \bibinfo{booktitle}{\emph{Advances in neural information processing
  systems}}. \bibinfo{pages}{5998--6008}.
\newblock


\bibitem[\protect\citeauthoryear{Wu, Schuster, Chen, Le, Norouzi, Macherey,
  Krikun, Cao, Gao, Macherey, et~al\mbox{.}}{Wu et~al\mbox{.}}{2016}]%
        {wu2016google}
\bibfield{author}{\bibinfo{person}{Yonghui Wu}, \bibinfo{person}{Mike
  Schuster}, \bibinfo{person}{Zhifeng Chen}, \bibinfo{person}{Quoc~V Le},
  \bibinfo{person}{Mohammad Norouzi}, \bibinfo{person}{Wolfgang Macherey},
  \bibinfo{person}{Maxim Krikun}, \bibinfo{person}{Yuan Cao},
  \bibinfo{person}{Qin Gao}, \bibinfo{person}{Klaus Macherey}, {et~al\mbox{.}}}
  \bibinfo{year}{2016}\natexlab{}.
\newblock \showarticletitle{Google's neural machine translation system:
  Bridging the gap between human and machine translation}.
\newblock \bibinfo{journal}{\emph{arXiv preprint arXiv:1609.08144}}
  (\bibinfo{year}{2016}).
\newblock


\bibitem[\protect\citeauthoryear{Wu, Xiong, Yu, and Lin}{Wu
  et~al\mbox{.}}{2018}]%
        {wu2018unsupervised}
\bibfield{author}{\bibinfo{person}{Zhirong Wu}, \bibinfo{person}{Yuanjun
  Xiong}, \bibinfo{person}{Stella~X Yu}, {and} \bibinfo{person}{Dahua Lin}.}
  \bibinfo{year}{2018}\natexlab{}.
\newblock \showarticletitle{Unsupervised feature learning via non-parametric
  instance discrimination}. In \bibinfo{booktitle}{\emph{Proceedings of the
  IEEE Conference on Computer Vision and Pattern Recognition}}.
  \bibinfo{pages}{3733--3742}.
\newblock


\bibitem[\protect\citeauthoryear{Xie, Lai, Doran, and Kadav}{Xie
  et~al\mbox{.}}{2019}]%
        {xie2019visual}
\bibfield{author}{\bibinfo{person}{Ning Xie}, \bibinfo{person}{Farley Lai},
  \bibinfo{person}{Derek Doran}, {and} \bibinfo{person}{Asim Kadav}.}
  \bibinfo{year}{2019}\natexlab{}.
\newblock \showarticletitle{Visual entailment: A novel task for fine-grained
  image understanding}.
\newblock \bibinfo{journal}{\emph{arXiv preprint arXiv:1901.06706}}
  (\bibinfo{year}{2019}).
\newblock


\bibitem[\protect\citeauthoryear{Yu, Yu, Cui, Tao, and Tian}{Yu
  et~al\mbox{.}}{2019}]%
        {yu2019deep}
\bibfield{author}{\bibinfo{person}{Zhou Yu}, \bibinfo{person}{Jun Yu},
  \bibinfo{person}{Yuhao Cui}, \bibinfo{person}{Dacheng Tao}, {and}
  \bibinfo{person}{Qi Tian}.} \bibinfo{year}{2019}\natexlab{}.
\newblock \showarticletitle{Deep modular co-attention networks for visual
  question answering}. In \bibinfo{booktitle}{\emph{Proceedings of the IEEE
  Conference on Computer Vision and Pattern Recognition}}.
  \bibinfo{pages}{6281--6290}.
\newblock


\bibitem[\protect\citeauthoryear{Zellers, Bisk, Farhadi, and Choi}{Zellers
  et~al\mbox{.}}{2019}]%
        {zellers2019recognition}
\bibfield{author}{\bibinfo{person}{Rowan Zellers}, \bibinfo{person}{Yonatan
  Bisk}, \bibinfo{person}{Ali Farhadi}, {and} \bibinfo{person}{Yejin Choi}.}
  \bibinfo{year}{2019}\natexlab{}.
\newblock \showarticletitle{From recognition to cognition: Visual commonsense
  reasoning}. In \bibinfo{booktitle}{\emph{Proceedings of the IEEE/CVF
  Conference on Computer Vision and Pattern Recognition}}.
  \bibinfo{pages}{6720--6731}.
\newblock


\bibitem[\protect\citeauthoryear{Zhu, Cheng, Gan, Sun, Goldstein, and Liu}{Zhu
  et~al\mbox{.}}{2020}]%
        {zhu2020freelb}
\bibfield{author}{\bibinfo{person}{Chen Zhu}, \bibinfo{person}{Yu Cheng},
  \bibinfo{person}{Zhe Gan}, \bibinfo{person}{Siqi Sun}, \bibinfo{person}{Tom
  Goldstein}, {and} \bibinfo{person}{Jingjing Liu}.}
  \bibinfo{year}{2020}\natexlab{}.
\newblock \showarticletitle{FreeLB: Enhanced Adversarial Training for Natural
  Language Understanding}. In \bibinfo{booktitle}{\emph{ICLR}}.
\newblock
\urldef\tempurl%
\url{https://openreview.net/forum?id=BygzbyHFvB}
\showURL{%
\tempurl}


\end{thebibliography}



\end{document}